\documentclass[a4paper,fleqn]{cas-sc}
\usepackage[numbers,longnamesfirst]{natbib}
\usepackage{subfigure}

\begin{document}
\let\WriteBookmarks\relax
\def\floatpagepagefraction{1}
\def\textpagefraction{.001}

\title[mode = title]{A Knowledge-based Learning Framework for Self-supervised Pre-training Towards Enhanced Recognition of Biomedical Microscopy Images}

\tnotemark[1]
% Short title
\shorttitle{A Knowledge-based Learning Framework for Self-supervised Pre-training Towards Enhanced Recognition of Biomedical Microscopy Images}    

% Short author
\shortauthors{Chen et. al.}  
% Funding
\tnotetext[1]{This work was supported by the National Key Research and Development Program of China (No. 2018YFB0204301) and the Natural Science Foundation of Hunan Province of China (No. 2022JJ30666).}

% Author
\author[1]{Wei Chen}

\address[1]{National University of Defense Technology, Changsha 410073, China}
\author[1]{Chen Li}
\cormark[1] 
\cortext[1]{Corresponding author} 
\author[2]{Dan Chen}
\author[1]{Xin Luo}
\address[2]{Wuhan University, Wuhan 430072, China}

\begin{abstract}
Self-supervised pre-training has become the priory choice to establish reliable neural networks for automated recognition of massive biomedical microscopy images, which are routinely annotation-free, without semantics, and without guarantee of quality.
Note that this paradigm is still at its infancy and limited by closely related open issues: 
1) how to learn robust representations in an unsupervised manner from unlabelled biomedical microscopy  images of low diversity in samples? and 
2) how to obtain the most significant representations demanded by a high-quality segmentation?
Aiming at these issues, this study proposes a knowledge-based learning framework (TOWER) towards enhanced recognition of biomedical microscopy images, which works in three phases by synergizing contrastive learning and generative learning methods:
1) Sample Space Diversification: Reconstructive proxy tasks have been enabled to embed \emph{a priori} knowledge with context highlighted to diversify the expanded sample space;
2) Enhanced Representation Learning: Informative noise-contrastive estimation loss regularizes the encoder to enhance representation learning of annotation-free images;
3) Correlated Optimization: Optimization operations in pre-training the encoder and the decoder have been correlated via image restoration from proxy tasks, targeting the need for semantic segmentation.
Experiments have been conducted on public datasets of biomedical microscopy images against the state-of-the-art counterparts (e.g., SimCLR and BYOL), and results demonstrate that: TOWER statistically excels in all self-supervised methods, achieving a Dice improvement of 1.38 percentage points over SimCLR. 
TOWER also has potential in multi-modality medical image analysis and enables label-efficient semi-supervised learning, e.g., reducing the annotation cost by up to 99\% in pathological classification.
\end{abstract}

% Research highlights
% \begin{highlights}
% 	\item Aiming at the grand challenges for automated recognition of biomedical microscopy  images towards clinical practices, this study developed a priori knowledge-based learning framework, which bridged generative and contrastive learning as a whole and provided better performance and faster convergence for segmentation by correlated-optimization between the encoder and decoder.  	
%     \item The proposed framework outperformed even the latest high-performance counterparts in terms of recognizing unlabelled biomedical microscopy  images. It suggested that the key to sustaining reliable recognition of unlabelled biomedical microscopy  images lied with appropriate portraying a priori knowledge in optimization.
%     \item To the best of our knowledge, the proposed framework is the first to explore the problem of insufficient diversity of contrastive learning for the recognition of biomedical microscopy  images.
%     \item The proposed framework could mitigate the lack of annotations, resulting in label-efficient representation learning for biomedical microscopy  image recognition (reduce up to 99\% annotations in pathological classification).
% \end{highlights}

\maketitle

% \linenumbers
% Main text
\section{Introduction}
\label{sec:introduction}
The success of nowadays biomedical research and clinical practices have largely relied on automated recognition of massive biomedical microscopy images, sustaining fine-grained interpretation of the physiological and pathological states of organs, tissues, and lesions. These images generated with biomedical microscopy techniques are routinely annotation-free, highly similar, and without guarantee of quality. It still remains an active research area to reach reliable conclusions based on the results of critical tasks such as classification and segmentation of these biomedical images, as routine end-to-end recognition neural networks largely rely on excessive labeling by human experts.  

Self-supervised learning has proved powerful in learning representations without the need for large labelled datasets~\cite{jing2020self}. Self-supervised pre-training has become essential in harsh scenarios like recognition of biomedical microscopy images to obtain state-of-the-art performance using unlabelled data~\cite{zoph2020rethinking}. Self-supervised pre-training routinely aims to pre-train an Auto-encoders (AE) neural network on a large amount of unlabelled images. It then adopts the well-trained neural network~\footnote{All mentions of the ``neural network" in this paper refer to Auto-encoders.} to downstream tasks with ``optimal" initialization ensured, possibly complemented with alternative networks when necessary.  

Cutting-edge methods of self-supervised pre-training targeting on image recognition are largely established on \emph{contrastive learning}~\cite{liu2021self}, which centers on how to ``learn to compare" to construct a high-quality representation space. Contrastive learning methods can be \emph{context-instance contrast} and \emph{instance-instance contrast}:
\begin{itemize}

\item Context-instance contrast models the mutual information (MI) between the local feature and its global context, and the representation space may be optimized by maximizing the MI. These approaches (e.g., InfoMax~\cite{hjelm2018learning}) can extract the most discriminative local representations for downstream classification tasks. Note that MI measurement is highly computing-intensive, and performance bottleneck needs to be tackled in this context.

\item Instance-instance contrast directly measures the similarity between different samples. It then extracts the instance-level representations by pulling the positive (similar) pairs together and pushing the negative(dissimilar) pairs apart. These approaches (e.g., SimCLR~\cite{chen2020simple} and MoCo~\cite{he2020momentum}) become dominant in classification tasks with performance competitive with supervised-based alternatives. However, the performance degrades when recognizing biomedical microscopy images due to insufficient sample diversity.
\end{itemize} 

Unfortunately, contrastive learning methods generally cannot suffice in dense prediction tasks, and the segmentation (via AE by default) of biomedical microscopy images is exactly the case. These tasks demand correlated optimization between the encoder and decoder, where contrastive learning is designed to optimize encoders only, leaving decoder training unattended.  

%Besides, unlabelled medical images routinely have low quality and tiny inter-class differences, resulting in insufficient sample diversity to support contrastive learning.

%Unlike contrastive learning that only optimizes encoders for classification tasks,
Generative learning is another self-supervised paradigm that learns the context-instance representations by restoring the original data distribution from transformations. It does not assume downstream tasks in advance, which can then provide fast and consistent initialization for classification and segmentation tasks. Note that its performance is not satisfied compared to contrastive learning when recognizing biomedical microscopy images.

Consequently, self-supervised pre-training is still at its infancy for biomedical microscopy images despite the success that has been achieved. When handling killer applications in scenarios as harsh as biomedical microscopy image recognition targeting at clinical practices, this paradigm is refrained by the closely related open issues:

How to learn robust representations in an unsupervised manner from unlabelled biomedical microscopy images of low diversity in samples? Unlabelled biomedical microscopy images are routinely with only insignificant inter-class differences. Insufficient sample diversity is a constant under this circumstance, while sufficient positive/negative pairs are mandatory for any successful contrastive learning (marked as Issue \#1).

How to obtain the most significant representations demanded by a high-quality segmentation? Segmentation as a dense prediction task demands collaboration between the encoder and the decoder. Contrastive learning is encoder-oriented only, while the performance of solutions based on generative learning is not satisfied (Issue \#2). 

This study first needs to extend the sample space. A priori knowledge of target tissues in biomedical microscopy images may help in enriching the stylistic and structural diversity of the sample space, while direct brute-force learning of the unlabelled images with insignificant differences does not apply. After that, contrastive learning may construct more diverse positive/negative pairs to extract instance-level representations, while generative learning is capable of learning context-level representations. Note that generative learning excels in the co-initialization of the auto-encoder. It is desirable to bridge generative and contrastive learning to co-optimize the encoder and decoder. Thus high-quality representations may be obtained for segmentation tasks by befitting from both methods' merits. 

Aiming at these issues, this study proposes a knowledge-based learning framework (\textbf{TOWER}) \textbf{TOW}ards \textbf{E}nhanced \textbf{R}ecognition of biomedical microscopy images, which works in three phases:

\begin{itemize}

\item  Sample Space Diversification (Section~\ref{ssec:diversification}): 
Reconstructive proxy tasks have been designed to perform the nonlinear translation and random masked reconstruction based on a priori knowledge from clinic practices. Transformed images are obtained via these tasks with the stylistic and structural diversity of sample space enriched. 

\item  Enhanced Representation Learning (Section~\ref{ssec:reinforcement}): The transformed images form the basis of constructing positive/negative sample pairs of higher diversity, which sustains the need for contrastive learning. Informative noise contrast estimation (InfoNCE) loss regularizes the feature space extracted by the encoder, which can enhance the representation learning of annotation-free images.

\item  Correlated Optimization (Section~\ref{ssec:optimization}): Generative learning makes full use of the powerful representations from the last phase and applies MSE loss to guide the optimization of the encoder-decoder, which reconstructs the transformed images and enhances the representation learning towards style and structural context. The correlated optimization then bridges contrastive and generative learning and serves the need for semantic segmentation.
\end{itemize}

Note that the proposed method defines the regions of interest (ROI) specified into the shape of different masks (Section~\ref{ssec:diversification}) characterizing typical biomedical microscopy images. For example, there is a strong physiological relationship between the optic disc and blood vessels in the retinas from fundoscopic images, so it can be characterized through a biological vision perspective. Specifically, the rays-wise mask applies to reconstruct the physiological relationships between the optic disc and blood vessels. The pre-trained neural network can extract the target tissue features in the ROI by masking these regions and reconstructing them afterwards, i.e., making use of the important a priori knowledge.

% \begin{itemize}
% \item For retinal images, the rays-wise mask applies to reconstruct the physiological relationships between the optic disc and blood vessels;
% \item For X-ray images, the stripe-wise mask applies to reconstruct the texture information of the bones distributed in the stripe region;
% \item For CT images, the block-wise mask applies to reconstruct the distribution of abdominal organs. 
% \end{itemize}

Extensive experiments have been performed on public biomedical microscopy image datasets (e.g., DRIVE) against the state-of-the-art counterparts (e.g., SimCLR and BYOL). TOWER's performance (AUC/Dice) and convergence have been evaluated. Supplementary tests have been made to examine the potentials of TOWER with other types of medical images (e.g., CT and X-ray). TOWER's label efficiency has been examined with different percentages of partially labelled images in a semi-supervised manner. 

The main contributions of this study are as follows:
\begin{itemize}
\item This study develops a knowledge-based learning framework to recognize biomedical microscopy images without annotations via self-supervised pre-training. TOWER significantly improves downstream tasks' performance with enhanced convergence and label efficiency. 
\item To the best of our knowledge, the proposed framework is the first to tackle the problem of insufficient diversity of contrastive learning for the recognition of biomedical microscopy images. 
\item A correlated optimization between encoder and decoder is proposed to provide significant representations for initializing decoder demanded by high-quality segmentation of biomedical microscopy images.

\end{itemize}

\section{Related Work}
\label{Related work}

Recognition of unlabelled images had attracted tremendous attentions in the machine learning community, and it remained an intriguing issue to learn robust representations without annotations. Studies undertaken for this purpose centering on pre-training Auto-encoders generally followed two directions: 1) optimizing the encoder via contrastive learning, and/or 2) co-optimizing the encoder and the decoder via generative learning. The most salient works along these directions were introduced as follows.

Chen et al. developed a simple yet effective \emph{Contrastive Learning} framework SimCLR v1~\cite{chen2020simple} and explored data enhancement strategies on two symmetric encoder-mlp branches for contrastive learning. 
Azizi et al. ~\cite{azizi2021big} developed the SimCLR into medical image classification. Different from natural image classification, the proposed multi-instance contrastive learning method (MICLe) constructed two crops from the images of the same patient as positive pairs. MICLe outperformed the ImageNet-based supervised baselines but did not explore thedense pixel prediction tasks.

He et al. proposed MoCo v1~\cite{he2020momentum}, which introduced a dynamic dictionary to store negative samples with no need for large batch size. 
Sowrirajan et al. adopted this idea and developed it into chest X-ray images, named MoCo\_CXR~\cite{MoCo_cxr}. It demonstrated that pre-training with contrastive learning on medical images was superior to natural image-based pre-training schemes for X-ray interpretation tasks.

Taher et al. explored the collaborative effectiveness in extracting representations from unlabeled medical images. They proposed a unified pre-training framework (CAiD)~\cite{taher2022caid} to unite contrastive and restorative learning for medical image pre-training. After that, Haghighi et al. updated it with adversarial learning and proposed DiRA~\cite{haghighi2022dira}. However, they ignored the insufficient diversity of medical images and still relied on traditional data augmentations (e.g., random horizontal flipping and gaussian blurring), failing to mine the implicit knowledge for biomedical microscopy images.

%The sub-optimal performance in dense prediction tasks might result from the absent initialization of the decoder. 
To initialize the encoder and the decoder at the same time, diverse proxy tasks had been proposed to aid generative learning for more effective pre-training via~\cite{Noroozi_2018_CVPR,Colorization,CHEN2019101539}. Attempts had been made along the direction of \emph{Generative Learning}:

Zhou et al. proposed a unified pre-training framework (Model Genesis~\cite{ZHOU2021101840}) for 3D medical images that integrated various proxy tasks to transform images with encoder-decoder initialization, including non-linear transformation, local shuffling, and in/out painting. By predicting the original images from the transformation, the framework enabled self-supervised representation learning for CT/MRI 3D image analysis. Note that generative learning methods generally were not able to compete with supervised pre-training counterparts when handling 2D medical images~\cite{zhou2019models}. 

He et al. proposed Masked AutoEncoder (MAE)~\cite{mae} and used block-wise masks in model training to reconstruct the randomly-masked input images. 
Chen et al. ~\cite{medmae} adopted this idea and advanced it into 3D medical image analysis. They used masked image modeling approaches to achieve faster convergence than supervised pre-training. However, the block-wise masking strategies were built on the ViT~\cite{dosovitskiy2021an} and might not directly apply to most biomedical microscopy image analysis, where ROI was not distributed in blocks and task-related semantic characteristics were not guaranteed. 

Inspired by the successes of the existing work, this study aimed at self-supervised pre-training towards enhanced recognition of biomedical microscopy images via the synergy of contrastive learning and generative learning: 1) to enrich the diversity of biomedical microscopy images with a priori knowledge, 2) to enhance self-supervised learning in terms of instance-level and context-level representations, and 3) to provide high-quality initialization for dense prediction tasks.
 
% \newpage
\section{Knowledge-based learning framework towards enhanced recognition of biomedical microscopy images}\label{sec:method}

This section first presents the overall design of TOWER and then details the working mechanism of TOWER in three aspects: 1) sample space diversification, 2) enhanced representation learning, and 3) correlated optimization.

\subsection{Overall Design}
\label{Framework}
\begin{figure*}[tb]
    \centering
    %\vspace{-0.5cm}
    \includegraphics[width=\textwidth]{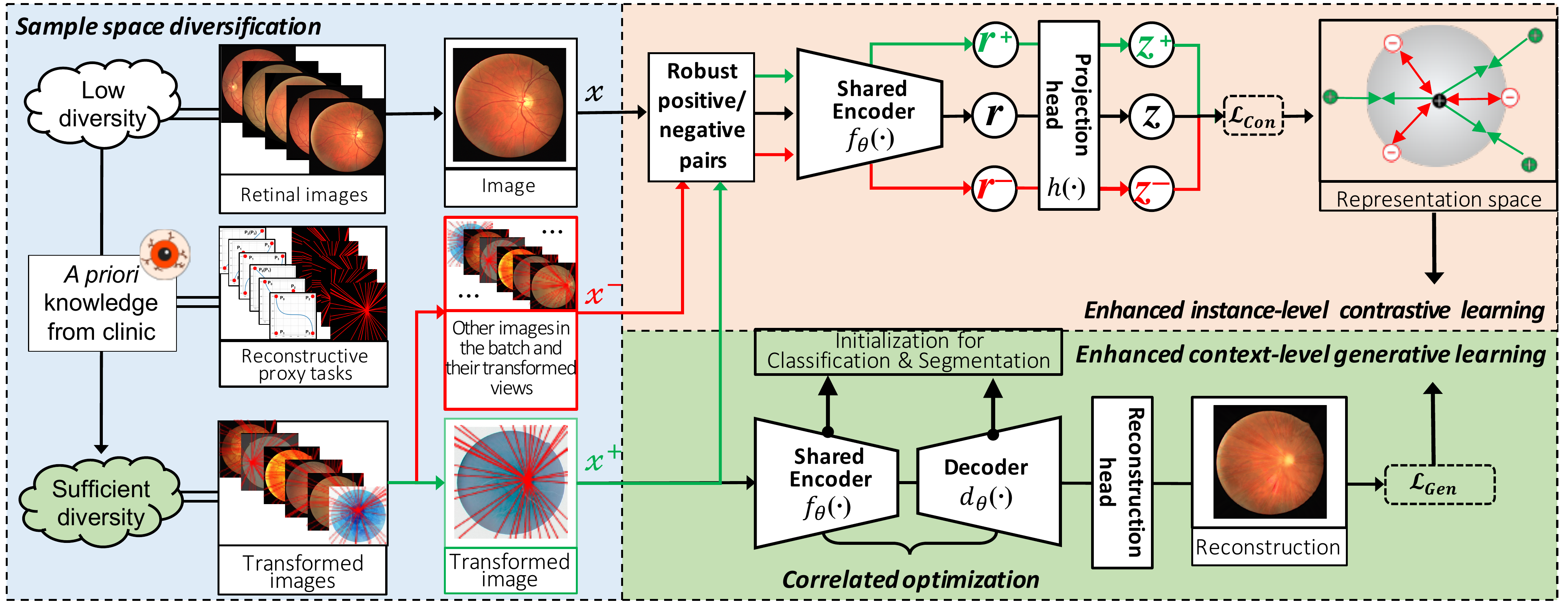}
    \caption{Overall design of TOWER framework. $\boldsymbol{r}$ and $\boldsymbol{z}$ respectively denote \emph{representations} and \emph{embeddings}. The green and red arrows respectively represent \emph{positive} and \emph{negative pairs}. Note that fundoscopic images are highlighted in the figures of this paper to illustrate the proposed framework.
    }
    %\vspace{-0.2cm}
    \label{fig:framework}
\end{figure*}

Fig.\ref{fig:framework} gives an overview of TOWER framework, where a 2D U-Net~\cite{UNET} is selected as the encoder-decoder with parameters $\theta$, denoted as $f_\theta(\cdot)$ and $d_\theta(\cdot)$. The backbone of the encoder $f_\theta(\cdot)$ is a ResNet-50~\cite{7780459}-based network with an MLP-based classification head $h(\cdot)$. The AE neural network receives an input $X\in\mathbb{R}^{N\times H\times W\times C}$, which is a randomly sampled batch with $N$ images; the output is the dense prediction $Y\in\mathbb{R}^{N\times H\times W\times C}$ with the same resolution as $X$. The objective of neural network training is to properly initialize both the encoder and decoder to serve the need of downstream tasks with high-quality representations. 

As a self-supervised learning framework, TOWER aims to optimize the $\theta$ from unlabelled retinal images $X$ so that $f_\theta(\cdot)$ and $d_\theta(\cdot)$ can be efficiently fine-tuned using only a few labelled examples when transferring to downstream tasks, e.g., classification and segmentation. 
% TOWER operates as follows (see Alg.~\ref{alg:Pseudocode} for pseudocode):

\begin{itemize}
    \item Reconstructive proxy tasks $\phi(\cdot)$ embed a priori knowledge from clinic practices. Knowledge-based nonlinear translation (Section~\ref{sssec:nonlinear}) and masked reconstruction (Section~\ref{sssec:mask}) are proposed to enrich the stylistic and structural diversity of sample space, respectively;
    
    \item Contrastive learning routine (based on SimCLR v1~\cite{chen2020simple}) constructs positive/negative sample pairs from the diversified sample space (Section~\ref{sssec:contrastive}). The InfoNCE~\cite{oord2019representation} loss function is used to regularize the encoder to learn instance-level representations. 

    \item
    Generative learning (enhanced Model Genesis~\cite{zhou2019models} in this study) restores the transformed images from reconstructive proxy tasks (Section~\ref{sssec:generative}). The MSE loss function is used to regularize the encoder and decoder to make consistent reconstructions and learn context-level representations.

    \item Contrastive and generative learning share the auto-encoder and the strengths between the two are complementary (Section~\ref{ssec:optimization}). Contrastive learning optimizes the encoder and provides more powerful representations for generative learning to restore stylistic and structural context. Generative learning optimizes the encoder and decoder, providing significant representations demanded by a high-quality segmentation for contrastive learning.
\end{itemize}

This design assumes that embedding a priori knowledge can offer an unrivalled opportunity for unsupervised representation learning of biomedical microscopy images. Contrastive learning and generative learning are bridged via reconstructive proxy tasks, and the two are then mutually enhanced for instance-level and context-level unsupervised representation. Correlated optimization of both the encoder and the decoder becomes possible with fine-tuned initialization to sustain tasks of biomedical microscopy image classification and dense prediction.

\subsection{Sample Space Diversification: a priori knowledge-embedded reconstructive proxy tasks}
\label{ssec:diversification}

\subsubsection{Enriching stylistic diversity via knowledge-based nonlinear translation}
\label{sssec:nonlinear}

In order to augment the biomedical microscopy images, a nonlinear translation proxy task has been designed. This task operates centering on nonlinear transformation, which can change pixel-wise values in an array according to a specific nonlinear mapping relationship. The design here aims to utilize its merit in extending the solution space of linear problems into a non-linear variant~\cite{nonlinear}. 

Medical images can be characterized by their special imaging mechanism, i.e., different intensity values in most medical images convey various implicit semantics. Such a priori knowledge has been embedded in the nonlinear translation proxy task.
% Taking CT images for example, different organs and tissues absorb X-rays differently, and thus doctors can roughly discriminate different regions via their CT values (Hounsfield Unit, HU). The HU values of livers are mostly in the range of [45,65], kidneys in the range of [20,40], and lung parenchyma in the range of [-850,-910]. The HU values of water are 0, and the HU values of air are -1000.
Generally speaking from the perspective of a medical image, changing its pixel values and transforming the overall style will alter the semantic mapping relationships. In this sense, nonlinear translation holds potential in enriching the stylistic diversity of the original sample space. TOWER designs multiple sets of monotonic invertible functions, which allow the values of each pixel to be restored after changing under given rules. In other words, this design enables invertible transformations of the image style. Bézier Curve\footnote{\url{https://pomax.github.io/bezierinfo}} is applied to generating the above functions:
\begin{equation}
    \text{Bézier}(P,n,t)= {\textstyle \sum_{i=0}^{n}} \binom{n}{i} {(1-t)}^{n-i} \cdot t^i \cdot P_i,
\end{equation}
where $P$ denotes the set of interpolation points $\{P_i|^{n}_{i=1}\}$, and $t$ is an independent variable in the range [0,1]. The Bézier curve then forms by interpolating the endpoints and the control points.

TOWER implements the nonlinear translation proxy task upon the cubic Bézier ($n$=3) as follows:
\begin{equation}
p^{\prime}=\text{Bézier}(\{P_0,P_1,P_2,P_3\},3,p)=P_0(1-p)^3+3P_1(1-p)^2p+3P_2(1-p) p^2+P_3p^3,
\end{equation}
where $p$ denotes the pixel-wise value in the normalized $x_n$, $p^{\prime}$ is the transformed value in the translated $x_n^{\prime}$. $P_0$, $P_3$ are endpoints and $P_1$, $P_2$ are control points.

\begin{figure}[t]
    \centering
    %\vspace{-0.5cm}
    \includegraphics[width=.75\linewidth]{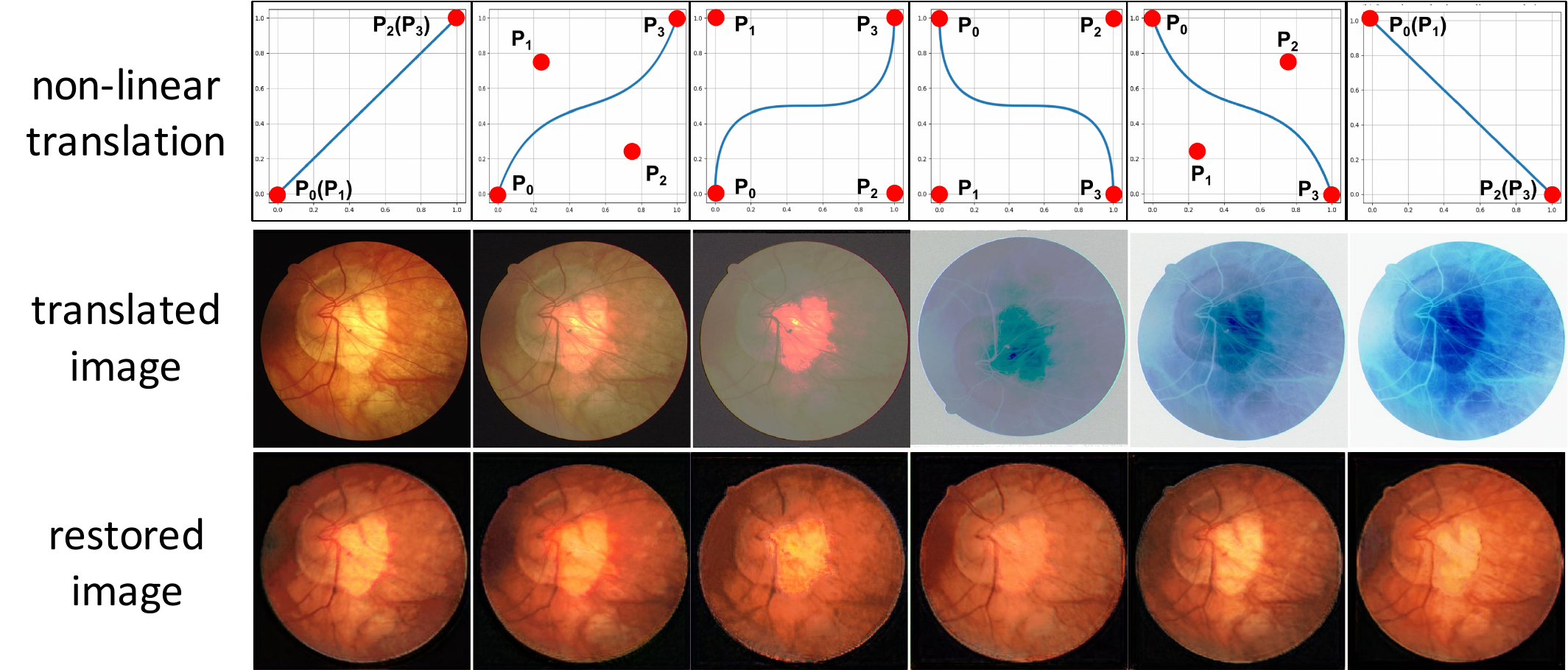}
    % %\vspace{-0.5cm}
    \caption{Illustrations of the nonlinear translations proxy task on fundoscopic images: six sets of Bézier curves-based functions (1st col) followed by translated images and corresponding reconstructed results.}
    %\vspace{-0.4cm}
    \label{fig:nonlinear}
\end{figure}

Fig.\ref{fig:nonlinear} illustrates the translation functions: (1) they increase monotonically when $P_0=(0,0)$ and $P_3=(1,1)$ (shown in the 1st, 2nd and 3rd rows), (2) they decrease monotonically when $P_0=(0,1)$ and $P_3=(1,0)$ (shown in the 4th, 5th and 6th rows). Note that the translation functions are linear (shown in the 1st and 6th rows) when $P_0=P_1$ and $P_2 = P_3$, respectively.

\subsubsection{Enriching structural diversity via knowledge-based masked reconstruction}
\label{sssec:mask}

Inspired by MAE~\cite{mae}, TOWER introduces randomly masked reconstruction as a proxy task to augment the translated images $x_n'$. A priori knowledge from the clinic is embedded into masks in this course, which considers the structure of common target tissues in biomedical microscopy images of different modalities. This design aims to enrich the structural diversity of the sample space.
\begin{figure}[t]
    \centering
    %\vspace{-0.5cm}
    \includegraphics[width=.75\linewidth]{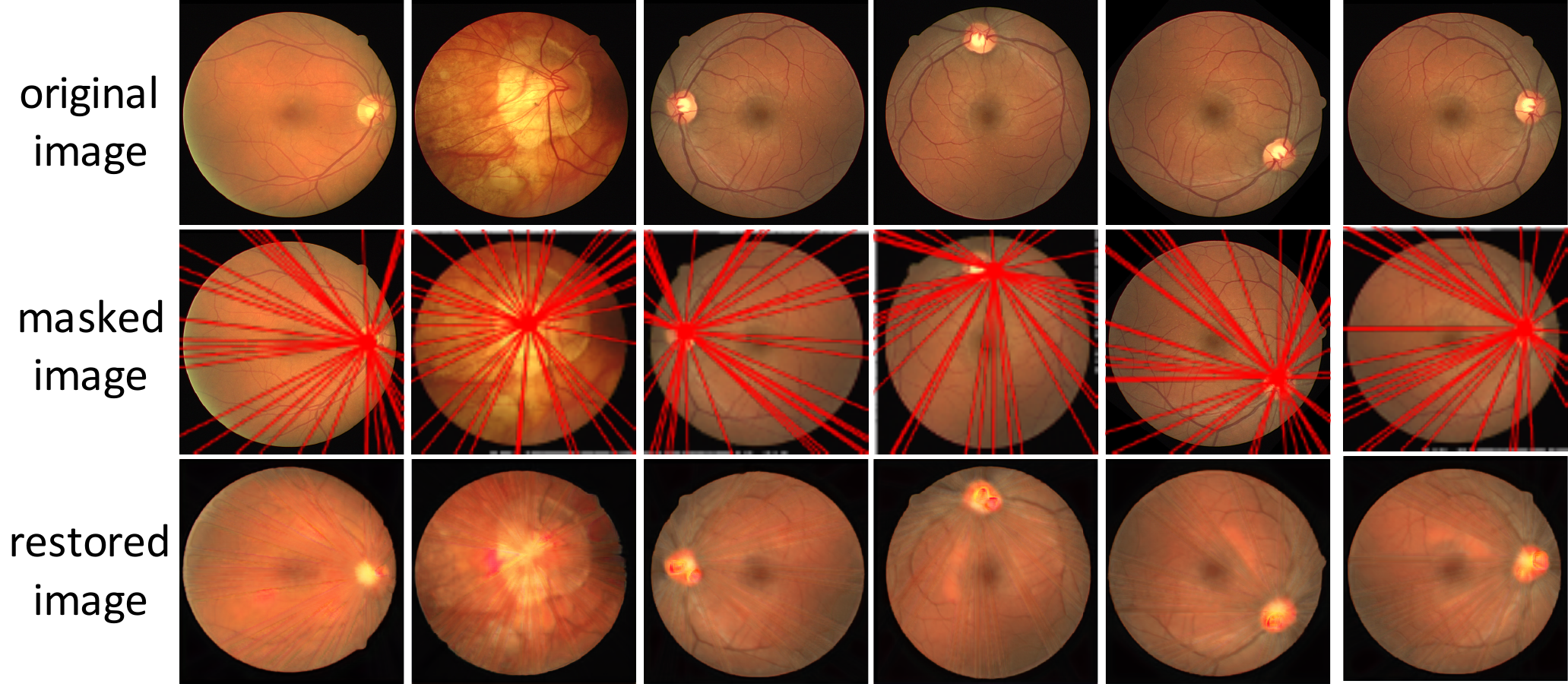}
    % %\vspace{-0.5cm}
    \caption{
    Illustration of the randomly masked reconstruction proxy task for fundoscopic images. Rays-wise masks are applied to augment the original images. TOWER reconstructs the structural information by predicting the pixel values for each masked pixel. TOWER can then restore the masked images and learn the structural context.}
    %\vspace{-0.2cm}
     \label{fig:mask}
\end{figure}

The regions of interest (ROI) are specified into the shape of the mask (see Fig.\ref{fig:mask} for the masks and the masked images). Taking fundoscopic images for example, the optic disc and blood vessels are the two most common tissues. They are the two most important ROIs in downstream tasks: 
\begin{itemize}
    \item The blood vessels of normal eyeballs are emitted from the optic disc. Analogizing the optic disc to a starting point, the vessels can be regarded as rays emanating from the starting point. The starting point of the rays-wise mask is located in the brightness point in the retina images. The proposed rays-wise mask can reasonably simulate the above physiological relationships;
    \item The diameter of the optic disc of normal people is about 1.5mm, while the diameter of the optic cup is approximately 1/3 of the optic disc, about 0.5mm. Moreover, the average diameter of blood vessels with uneven thickness is about 0.1mm. The mask can then be set according to the rays' thickness and the starting point's diameter.
\end{itemize}

\begin{figure}[tb]
    \centering
    % %\vspace{-0.5cm}
    \includegraphics[width=.6\linewidth]{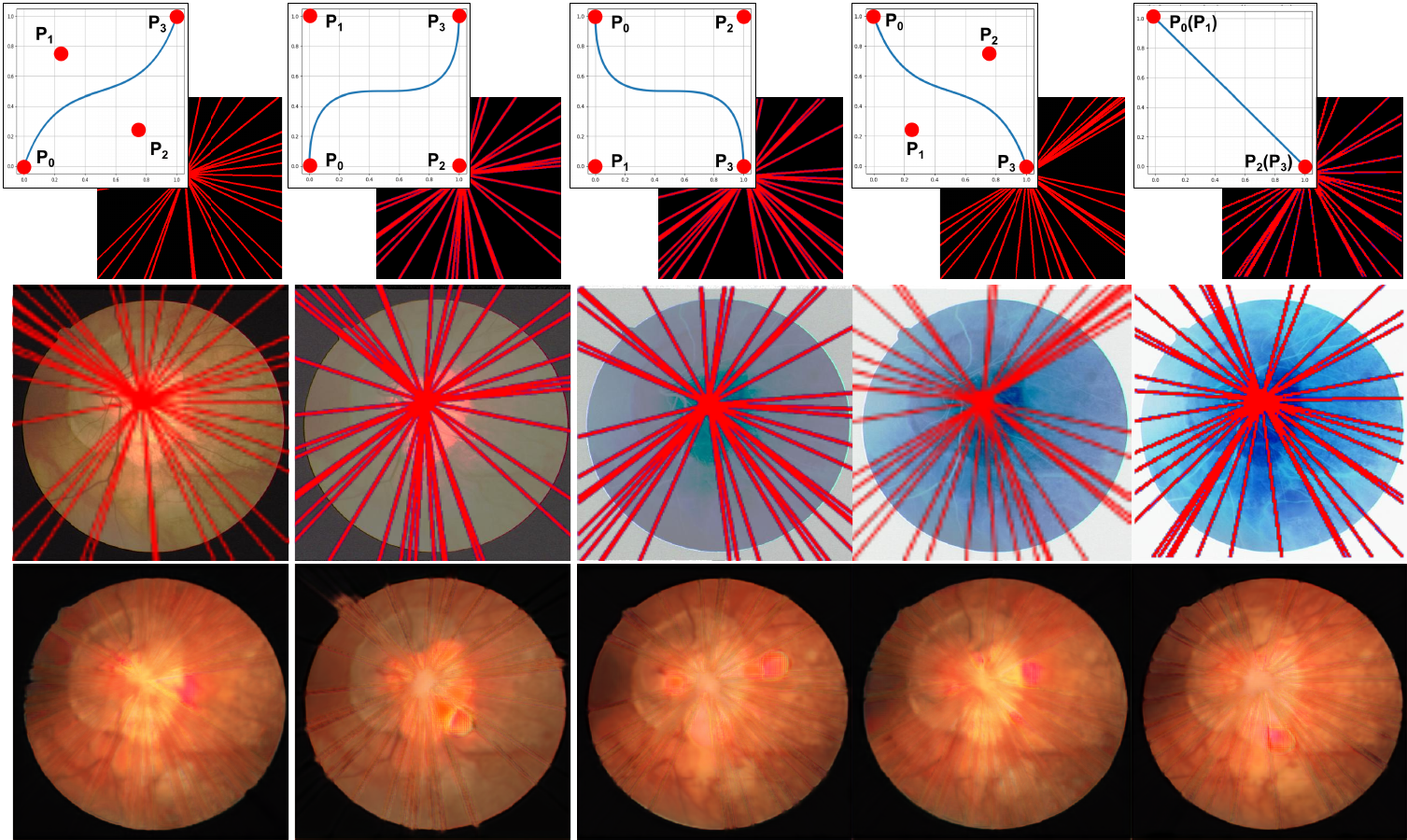}
    \caption{Reconstructive proxy task integrating a hybrid training scheme: The first rows represent the hybrid schemes; The second rows represent the transformed images; and the last rows represent the reconstruction. TOWER can learn the style and structure semantics.}
    %\vspace{-0.3cm}
    \label{fig:hybrid}
\end{figure}

Consequently, for fundoscopic images, the rays-wise mask reconstructs the physiological relationships between the optic disc and blood vessels.
After obtaining the mask $Mask$, the reconstruction proxy task obtains the transformed images $x_n^{\prime\prime} = x_n^{\prime}\times Mask$ via multiplying $x_n^{\prime}$ with $Mask$ pixel by pixel. TOWER extracts the context of the target tissue in the ROI by masking these regions and reconstructing them.
The two sets of training schemes, i.e., (1) random mask reconstructions and (2) nonlinear translation, are used in a hybrid manner in this study (Fig.\ref{fig:hybrid}).

\subsection{Enhanced unsupervised representation learning}
\label{ssec:reinforcement}

Given the image batch $X=\{x_1, x_2, ...,x_n, ..., x_N\}$ and its transformed views $X^{\prime\prime} =\{x_1^{\prime\prime}, x_2^{\prime\prime}, ..., x_n^{\prime\prime}, ..., x_N^{\prime\prime}\}$, TOWER bridges the contrastive and generative learning via reconstructive proxy tasks. It then enhances the instance-level and context-level unsupervised representation learning towards recognition of biomedical microscopy images in two complementary aspects.

\subsubsection{Instance-level representation learning via contrastive learning}
\label{sssec:contrastive}

\begin{figure}[tb]
    \centering
    %\vspace{-0.5cm}
    \includegraphics[width=.8\textwidth]{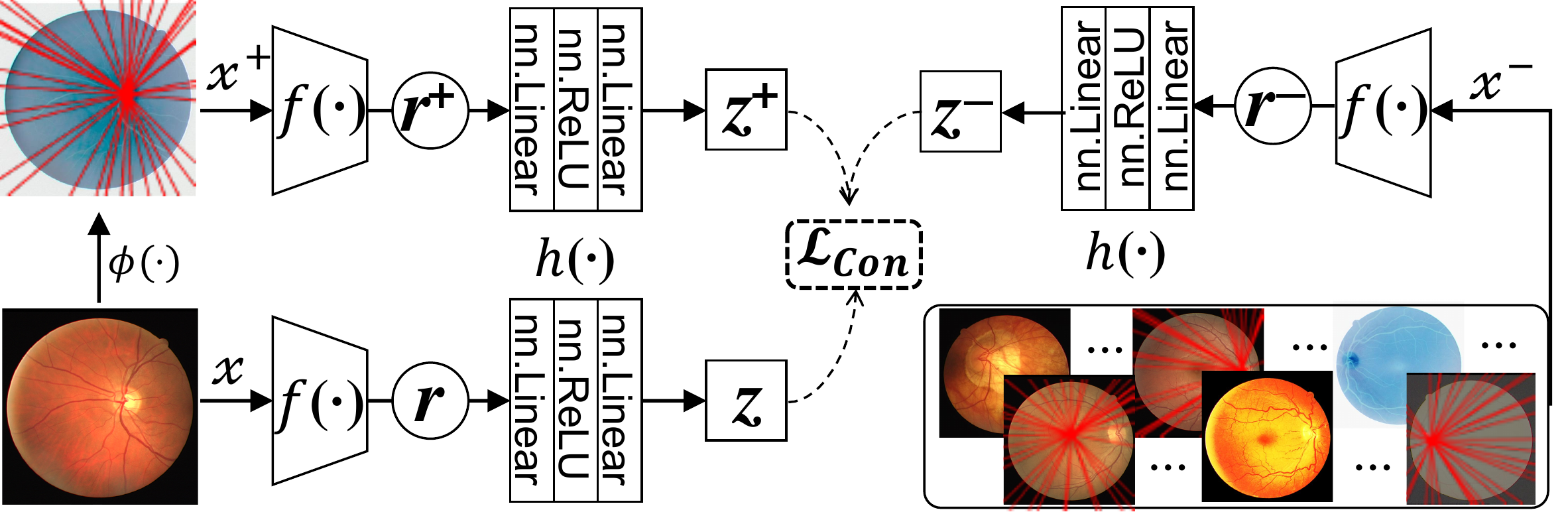}
    \caption{The contrastive learning workflow. 
    One sample and its transformed views by reconstructive proxy tasks are defined as positive pairs. The remaining samples in this batch and their transformed views are negative pairs.
    }%\vspace{-0.5cm}
    \label{fig:constrast}
\end{figure}

TOWER customizes the contrastive learning (SimCLR~\cite{chen2020simple}) workflow on the diversified sample space as shown in Fig.\ref{fig:constrast}.  TOWER defines the sample $x_n$ and its transformed views $x_n^{\prime\prime}$ as positive pairs: ${x_n}^+$. The remaining samples and their transformed views are the negative pairs of $x_n$: $x_n^-$.

% Specifically, the sample $x$ and its positive/negative pairs (${x}^+$,$x^-$) are 
For the sample and its positive/negative pairs ($x_n$/${x_n}^+$/$x_n^-$), a base encoder $\boldsymbol{f}_\theta(\cdot)$ extracts the representations ($\boldsymbol{r_n}$/$\boldsymbol{r_n^+}$/ $\boldsymbol{r_n^-}$). 
An MLP as the projection head $\boldsymbol{h}(\cdot)$ maps the representations to embeddings ($\boldsymbol{z_n}$/$\boldsymbol{z_n^+}$/$\boldsymbol{z_n^-}$).

The cosine similarity function $sim(a,b)$ then measures the similarity between pairwise embeddings via the dot product between $\ell_2$ normalized $a$ and $b$:
\begin{equation}
    sim(a, b)=a^\intercal b/(\left \|a\right \|\left \| b\right \|),
\end{equation}

Finally, InfoNCE~\cite{oord2019representation} as the contrastive loss function $\mathcal L_{Con}$ applies to (1) pull the representations of positive pairs together and (2) push the representations of negative pairs apart, defined as follows:
\begin{equation}
    \mathcal L_{Con}= \frac{-1}{N}\sum_{n=1}^{N}\log \frac{e^{\left[sim\left(z_n, z_n^{+}\right) / \tau\right]}}{e^ {\left[sim\left(z_n, z_n^{+}\right) / \tau\right]}+\sum_{z_n^{-}}  e^ {\left[sim\left(z_n, z_n^{-}\right) / \tau\right]}}, 
\end{equation}
where $\tau>0$ is a scalar temperature (set as 0.1~\cite{chen2020simple}).

The representation space of the encoder can then be optimized by minimizing the distances of $(\boldsymbol{z_n},\boldsymbol{z_n^+})$ and maximizing the distances of $(\boldsymbol{z_n},\boldsymbol{z_n^-})$.

\subsubsection{Context-level representation learning via generative learning}
\label{sssec:generative}

TOWER enhances the generative learning (Model Genesis~\cite{zhou2019models}) workflow by restoring the transformed images from more diverse transformations, i.e., nonlinear translation and random masking (Section~\ref{ssec:diversification}).

The 2D U-Net~\cite{UNET} (an encoder-decoder architecture) makes the dense prediction $y_n=d_\theta(f_\theta(x_n^{\prime\prime}))$ based on the transformed images $X^{\prime\prime}$.
The masked regions of $X^{\prime\prime}$ evolves invisibly in the training process. The control points set $\{P_i|^{n}_{i=1}\}$ also evolves with different Bézier Curves generated for nonlinear translations.

Note that such a hybrid transforming scheme ensures that the $X^{\prime\prime}$ can not be reconstructed by fitting an interpolation function. Reconstructing biomedical microscopy images from these transformations aids learning context-level representations.
For example, reconstructing the optic disc and blood vessels from partially masked fundoscopic images contributes to learning the local context of these tissues. Restoring the correct values from style-translated images contributes to learning the global context of whole images.

Generative learning restores the transformed image by optimizing the following loss function.
\begin{equation}
    \mathcal L_{Gen} =  {\textstyle \sum_{n=1}^{N}} \ell_{mse}(x_n,y_n),
\end{equation}
where $\ell_{mse}$ is the mean squared error (MSE) function; The objective is 1) to keep $y_n$ the same as the original image $x_n$, and 2) to ensure the encoder-decoder learning the context representations.

\subsection{Correlated optimization between the encoder and decoder}
\label{ssec:optimization}

Contrastive learning and generative learning routines excel respectively in extracting instance-level and context-level representations. It is desirable to complement the two with each other to enhance the unsupervised representation learning process. 

Contrastive learning specializes in optimizing the encoder, which may provide more powerful representations for generative learning to restore stylistic and structural context.
Generative learning's merit in optimizing the encoder and the decoder can provide representations significant enough for contrastive learning to sustain dense prediction tasks, i.e., high-quality segmentations.
%throws away the projection head $\boldsymbol{h}(\cdot)$ and 
On completion of training, TOWER opts for the encoder $\boldsymbol{f}_\theta(\cdot)$ to initialize classification tasks. 
The encoder $\boldsymbol{f}_\theta(\cdot)$ and decoder $\boldsymbol{d}_\theta(\cdot)$ are applied for initialization in the segmentation tasks.

In summary, the design enables correlated optimization of both the encoder and the decoder, which are targeted on classification and segmentation in downstream tasks. 
\section{Experiments and Results}
\label{sec:experiment}

Experiments on biomedical microscopy images were conducted (1) to evaluate TOWER’s performance to recognize fundoscopic images in comparison with the state-of-the-art counterparts (Section~\ref{ssec:comparison}), (2) to validate the effectiveness of TOWER via ablation studies based on biomedical microscopy images classification and segmentation (Section~\ref{ssec:ablation}), and (3) to evaluate the semantic-consistency between the encoder and the decoder of TOWER (Section~\ref{ssec:discussion}).

\subsection{Datasets and Experiment settings}\label{ssec:setup}
Biomedical microscopy images of various resolutions and modalities were used in this section, including fundoscopic images (DRIVE, RetinaMNIST), pathological images (PathMNIST) and dermatoscopic images (DermaMNIST), to evaluate the effectiveness of TOWER. 

The DRIVE came from the Digital Retinal Images for Vessel Extraction~\footnote{\url{http://www.isi.uu.nl/Research/Databases/DRIVE}} challenge, which was obtained from a diabetic retinopathy screening project in the Netherlands. The screening population consisted of 400 diabetic subjects aged 25–90. DRIVE included 40 randomly selected fundoscopic images, 33 retinas of healthy people and 7 retinas with early mild diabetic retinopathy. The input size was resized as 512$\times$512. The official dataset contained the training set (20 images) and the test set (20 images). The mean Dice score was utilized as the metric for the segmentation of blood vessels, with the manual segmentation of the vessel as ground truth.

RetinaMNIST, PathMNIST and DermaMNIST came from the MedMNIST decathlon~\cite{medmnistv1,medmnistv2}, which was a lightweight AutoML benchmark for medical image classification~\footnote{\url{https://medmnist.com/}} and covered diverse data modalities, dataset scales, and tasks. 
The RetinaMNIST was collected from DeepDRiD~\footnote{\url{https://isbi.deepdr.org/data.html}}, consisting of 1600 retina fundus images. RetinaMNIST aimed to perform classification of 5-level grading of diabetic retinopathy severity. The source images of 3$\times$1736$\times$1824 are center-cropped and resized into 3$\times$28$\times$28.
The PathMNIST was collected from NCT-CRC-HE-100K~\cite{pathmnist}, consisting of 100K colorectal cancer pathological images for training and 7K images from a different clinical center for testing. This dataset was designed for a multi-class classification task, where 9 kinds of tissues were expected to classify. The source images of 3$\times$224$\times$224 are resized into 3$\times$28$\times$28.
The DermaMNIST was collected from HAM10000 ~\cite{tschandl2018ham10000,codella2019skin}, a large collection of multi-source dermatoscopic images of common pigmented skin lesions. The dataset consists of 10,015 dermatoscopic images categorized as 7 different diseases, formulized as a multi-class classification task. The source images of 3$\times$600$\times$450 are resized into 3$\times$28$\times$28.
Data augmentations included random rotation, gaussian noise, color dithering, and horizontal, vertical, and diagonal flipping. All images were normalized.

For the classification tasks, TOWER was pre-trained with cross-entropy loss. The batch size was 128. The initial learning rate of pre-training was $1e^{-3}$ and it decreased to $2e^{-4}$ for fine-tuning. The fine-tuning process ended after 100 epochs or early stopped with patience 30. 
For the segmentation tasks, the loss function was standard pixel-wise cross-entropy loss. The batch size was 32. The initial learning rate of pre-training was $1e^{-2}$ and it decreased to $1e^{-3}$ for fine-tuning. The fine-tuning process ended after 200 epochs or early stopped with patience 30. 

The run-time infrastructure for the experiments was mainly formed by PyTorch 1.10.0 with CUDA 10.2 over Four NVIDIA 1080Ti GPUs. The Adam optimizer~\cite{kingma2015adam} and cosine learning rate decay schedulers were applied. The performance was measured in terms of AUC (\%) and Dice (\%) for classification and segmentation, respectively. All results were evaluated without post-processing and reported in (mean$\pm$std.) across ten independent trials.

\subsection{Comparison with the state-of-the-art methods on a self-supervised benchmark}\label{ssec:comparison}
This study conducted comparison experiments with the state-of-the-art methods on a self-supervised benchmark~\cite{Benchmark}. 
TOWER was compared with InsDis~\cite{wu2018unsupervised}, Model Genesis~\cite{zhou2019models}, CMC~\cite{tian2020contrastive}, MoCo v1-v2~\cite{he2020momentum,chen2020improved}, SimCLR v1-v2~\cite{chen2020simple,NEURIPS2020_fcbc95cc}, PIRL~\cite{misra2020self}, PCL v1-v2~\cite{li2021prototypical},  SeLa v2~\cite{YM.2020Self-labelling,NEURIPS2020_70feb62b}, InfoMin~\cite{NEURIPS2020_4c2e5eaa}, BYOL~\cite{NEURIPS2020_f3ada80d}, DeepCluster v2~\cite{caron2018deep,NEURIPS2020_70feb62b}, SwAV~\cite{NEURIPS2020_70feb62b}, Barlow Twins~\cite{pmlr-v139-zbontar21a}, ImageNet-based supervised initialization, and random initialization.
The neural network was first initialized by these pre-training methods and then fine-tuned on fundoscopic images from the DRIVE dataset for blood vessel segmentation.

Table~\ref{tab:sotaresults-vfs} reported the results and the statistical analysis between TOWER and other methods, which indicated that TOWER could achieve significantly higher accuracy in the blood vessels segmentation task, demonstrating the robustness of TOWER for promising initialization in dense prediction tasks.
Specifically, the backbone of neural network initialized by TOWER achieved 0.72 points increase in the Dice score with p-value\textless0.001, compared with the second-best method (DeepCluster-v2).
In addition, Table~\ref{tab:sotaresults-vfs} indicated that the performances of most classic contrastive learning methods were inferior to the performances of the ImageNet-based supervised pre-training method for biomedical microscopy image recognition, such as MoCo-v1 and SimCLR-v1.
\begin{table}[tb]
  \centering
  %\vspace{-0.5cm}
   \caption{TOWER outperformed state-of-the-art self-supervised pre-training methods and ImageNet initialization on blood vessels segmentation task (DRIVE). Results style: \textbf{best}, \underline{methods that outperformed ImageNet-based initialization}.
%   We also report the p-values between the best and the second best results for each ratio to demonstrate the significance of TOWER.
   }%\vspace{-0.2cm}

    \begin{tabular}{c|c|c}
    \hline
    Pre-training &\multirow{1}[2]{*}{Venue} &Segmentation task (Dice) \\  
methods &  & DRIVE \\ \hline
Random init. &- & 78.27$\pm$0.40$^{\star\star\star}$ \\
ImageNet init. &-  & 79.20$\pm$0.34$^{\star\star\star}$ \\ \hline
InsDis & \texttt{CVPR"18}  & 79.03$\pm$0.34$^{\star\star\star}$ \\
Model Genesis & \texttt{MedIA"20}& \underline{79.22$\pm$0.30}$^{\star\star\star}$ \\
CMC & \texttt{ECCV"20} & \underline{79.50$\pm$0.45}$^{\star\star}$ \\
MoCo-v1 & \texttt{CVPR"20}  & 78.98$\pm$0.45$^{\star\star\star}$ \\
PIRL & \texttt{CVPR"20}&   \underline{79.24$\pm$0.42}$^{\star\star\star}$ \\
SimCLR-v1 & \texttt{ICML"20}&   79.00$\pm$0.18$^{\star\star\star}$ \\
MoCo-v2 & \texttt{arXiv"20} &  \underline{79.23$\pm$0.19}$^{\star\star\star}$ \\
SimCLR-v2 & \texttt{NeurIPS"20}& 78.72$\pm$0.37$^{\star\star\star}$ \\
SeLa-v2 & \texttt{ICLR"20}&   \underline{79.65$\pm$0.19}$^{\star\star}$ \\
InfoMin & \texttt{NeurIPS"20}&   \underline{79.63$\pm$0.30}$^{\star\star}$ \\
BYOL & \texttt{NeurIPS"20}&  \underline{79.39$\pm$0.22}$^{\star\star\star}$ \\
DeepCluster-v2& \texttt{ICLR"20} & \underline{79.66$\pm$0.21}$^{\star\star}$\\
SwAV & \texttt{NeurIPS"20}& \underline{79.65$\pm$0.14}$^{\star\star}$ \\
PCL-v1 & \texttt{ICLR"21}&   78.99$\pm$0.21$^{\star\star\star}$ \\
PCL-v2 & \texttt{ICLR"21}&   79.06$\pm$0.19$^{\star\star\star}$ \\
Barlow Twins & \texttt{ICML"21}&  \underline{79.48$\pm$0.16}$^{\star\star\star}$ \\\hline 
TOWER & \texttt{Ours}&  \textbf{80.38$\pm$0.41} \\ \hline
    \end{tabular}%
    
    \resizebox{\linewidth}{!}{
    \begin{tabular}{ccc}
\multicolumn{3}{l}{$\ddagger$ Reproduced by ourselves under the same protocol.} \\
\multicolumn{3}{p{.9\linewidth}}{$\star$ Statistical analysis between the compared method and our proposed TOWER was also conducted in each target task, where $^{\star}$ denoted TOWER significantly outperformed the method with p-value\textless0.005. $^{\star\star}$ denoted p-value\textless0.001. $^{\star\star\star}$ denoted p-value\textless0.0001.}
    \end{tabular}}
    
  \label{tab:sotaresults-vfs}%
    %\vspace{-0.5cm}
\end{table}%

\subsection{Ablation study}\label{ssec:ablation}
\begin{table*}[tbp]
  \centering
  %\vspace{-0.5cm}
  \caption{Performance of the proposed self-supervised pre-training components on various downstream tasks. Results style: \textbf{best}.
%   The significance test is reported in Table \ref{tab:p-value} and \ref{tab:p-value-2}. 
%   Symbol * indicates statistically significant improvement given by a Wilcoxon signed-rank test with $p\leq0.05$.
  }%\vspace{-0.4cm}
	\begin{center}
% 	\resizebox{\linewidth}{!}{
    \begin{tabular}{c|cccc|ccc|c}
    \hline
%     \multirow{1}[3]{*}{Methods} & \multirow{1}[2]{*}{$\mathcal L_{Con}^{moco}$} & \multirow{1}[3]{*}{$\mathcal L_{Gen}^{NL}$} & \multirow{1}[2]{*}{$\mathcal L_{Gen}^{M}$} & \multirow{1}[3]{*}{$\mathcal L_{Con}^{NL+M}$} & PathMNIST & BreastMNIST & DermaMNIST & RetinaMNIST & OrganMNIST & DRIVE \\
% \cline{6-11}          &  &  &  &  & \multicolumn{6}{c}{AUC or Dice (\%) $\uparrow$} \\
%     \hline
% \cline{6-11}          &  &  &  &  & \multicolumn{6}{c}{AUC or Dice (\%) $\uparrow$} \\
%     \hline
    \multirow{1}[2]{*}{Methods} & \multirow{1}[2]{*}{$\mathcal L_{Con}^{moco}$} & \multirow{1}[2]{*}{$\mathcal L_{Gen}^{NL}$} & \multirow{1}[2]{*}{$\mathcal L_{Gen}^{M}$} & \multirow{1}[2]{*}{$\mathcal L_{Con}^{NL+M}$} & \multicolumn{3}{c|}{Classification AUC ($\%$)} & Segmentation Dice ($\%$) \\
    &  &  &  & & RetinaMNIST& PathMNIST& DermaMNIST  & DRIVE  \\
\hline
    Random &  &  &  &   & 69.46$\pm$1.90 & 94.61$\pm$0.40 & 87.23$\pm$1.62 & 78.27$\pm$0.40 \\
    ImageNet &  &  &  &  & 71.96$\pm$1.07& 97.85$\pm$0.45 & 90.55$\pm$0.62 & 79.20$\pm$0.34 \\
    MoCo  & $\checkmark$ &  &  &  & 69.98$\pm$1.89 & 96.93$\pm$0.66 & 90.04$\pm$0.42 & 79.23$\pm$0.19 \\
    \hline
    \multicolumn{1}{c|}{\multirow{3}[5]{*}{TOWER}} &  & $\checkmark$ &  &  & 71.52$\pm$1.22 & 97.46$\pm$0.57 & 90.49$\pm$0.93 & 79.82$\pm$0.39 \\
          &  &  & $\checkmark$ &  & 71.20$\pm$1.34 & 97.79$\pm$0.54 & 90.13$\pm$0.68 & 79.60$\pm$0.54 \\
          &  & $\checkmark$ & $\checkmark$ &  & 71.91$\pm$1.30 &  97.83$\pm$0.38 & 90.90$\pm$0.47 & 79.94$\pm$1.16 \\
          & &  &  & $\checkmark$  &\underline{72.61$\pm$0.58} & \underline{98.03$\pm$0.27} & \underline{91.72$\pm$0.95} &\underline{80.31$\pm$0.45} \\
          &  &$\checkmark$ & $\checkmark$ & $\checkmark$ & \underline{\underline{\textbf{73.01$\pm$0.94}}} & \underline{\underline{\textbf{98.49$\pm$0.24}}} & \underline{\underline{\textbf{91.78$\pm$0.30}}} & \textbf{80.38$\pm$0.30} \\
    \hline
    \multicolumn{9}{p{\linewidth}}{Note: $\mathcal L_{Con}^{moco}$ denoted the classic contrastive learning method (MoCo) without the proposed proxy task. $\mathcal L_{Gen}^{NL}$ denoted the generative learning method with nonlinear translation. $\mathcal L_{Gen}^{M}$ denoted the generative learning method with random masked reconstruction. $\mathcal L_{Con}^{NL+M}$ denoted the contrastive learning method with the proposed knowledge-based reconstructive proxy tasks. 
    The \underline{results} represented that $\mathcal L_{Con}^{NL+M}$ was statistically significantly better than the $\mathcal L_{Con}^{moco}$ with p-value \textless 0.05. The \underline{\underline{results}} denoted TOWER was statistically significantly better than the $\mathcal L_{Gen}^{NL+M}$ with p-value\textless 0.05.}\\
    \end{tabular}%
    % }
    \end{center}
    %\vspace{-0.8cm}
  \label{tab:ablation-vfs}%
\end{table*}%
Comprehensive ablation studies had been conducted to evaluate the design of individual components, basically via biomedical microscopy image classification (RetinaMNIST, PathMNIST, DermaMNIST) and segmentation (DRIVE).  
\begin{figure}[tb]
    \centering
    % \vspace{-0.6cm}
    \subfigure{% on Shenzhen X-ray dataset
        \begin{minipage}{0.4\textwidth}
    % %\vspace{-0.6cm}
        \centerline{\includegraphics[width=\textwidth]{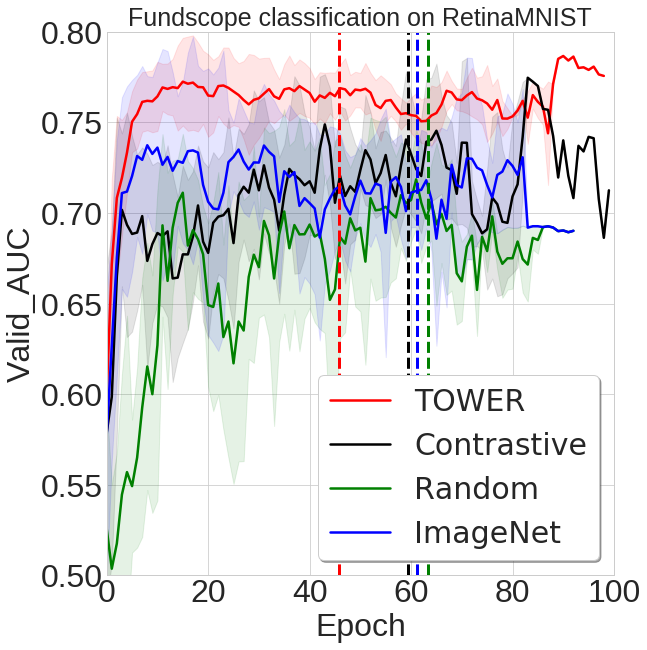}%
        }
    \end{minipage}
    }
    \subfigure{% on Montgomery X-ray dataset
        \begin{minipage}{0.4\textwidth}
    % %\vspace{-0.6cm}
        \centerline{\includegraphics[width=\textwidth]{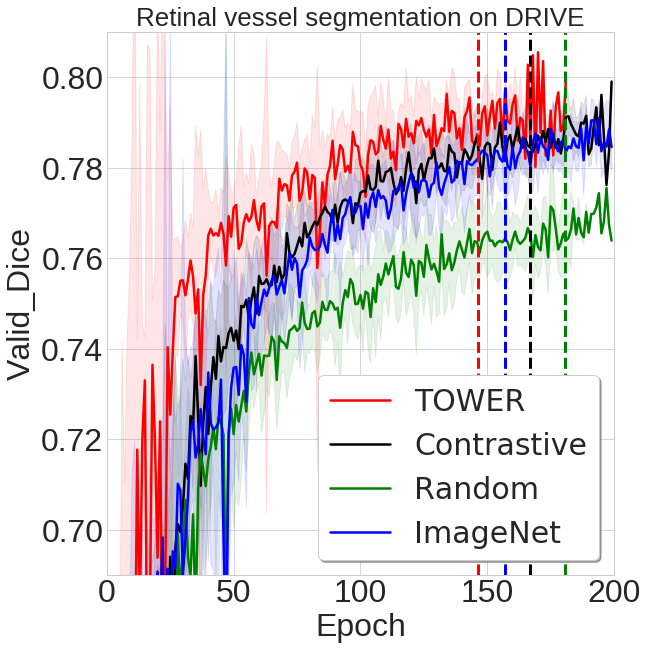}%
        }
        \end{minipage}
    }
    \caption{The convergence curve during validation. The vertical red, blue, black, and green dash lines indicated the early-stopped average epochs of TOWER, ImageNet, SimCLR v1, and random initialized.}
    %\vspace{-0.5cm}
    \label{fig:convergence}
\end{figure}

Experiments compared (1) the contrastive learning component with the proposed reconstructive proxy tasks (denoted as $\mathcal L_{Con}^{NL+M}$) and (2) MoCo with classic data augmentations such as random rotation, gaussian noise, color dithering, and horizontal, vertical, and diagonal flipping (denoted as $\mathcal L_{Con}^{moco}$). Table \ref{tab:ablation-vfs} reported the performance between the individual components and the combined scheme. Fig.~\ref{fig:convergence} showed the convergence curve during validation.
Experimental results indicated that:
\begin{itemize}
    \item When applying the classic contrastive learning method (MoCo) to recognize biomedical microscopy images, we observed that the $\mathcal L_{Con}^{moco}$ did not perform as well as pre-trained ones from ImageNet in most studies. In contrast, the proposed contrastive learning component outperformed the classic one, i.e., the $\mathcal L_{Con}^{NL+M}$ outperformed the $\mathcal L_{Con}^{moco}$ with p-value\textless0.05 on all listed tasks, demonstrating the effectiveness of reconstructive proxy tasks in enhancing contrastive learning (Issue \#1).
    \item The results of the combined training scheme outperformed any single reconstructive proxy task, demonstrating the scalability and effectiveness of the proposed reconstructive proxy tasks in enhancing generative learning.
    \item The whole framework achieved the best performance and the fastest convergence of all components after synergizing contrastive and generative learning, demonstrating the effectiveness of complementary training of generative and contrastive learning (Issue \#1).
\end{itemize}

\subsection{Comparison with different encoder-decoder-oriented pre-training schemes}\label{sssec:finetuning}
In order to investigate the semantic-consistency of different encoder-decoder-oriented pre-training schemes on downstream tasks, this section conducted fine-tuning experiments with the following initialization schemes on the DRIVE dataset:
(1) Random initialization for encoder and decoder,
(2) ImageNet-based pre-training of encoder~\footnote{Checkpoint was provided by \texttt{ torchvision.models.resnet50 (pretrained=True)}.},
(3) Generative learning-based pre-training of encoder and decoder~\footnote{Generative learning pre-trained a U-Net as encoder-decoder via loss function $\mathcal L_{Gen}^{NL}$.},
(4) Contrastive learning-based pre-training of encoder~\footnote{Contrastive learning pre-trained a ResNet50 as encoder via loss function $\mathcal L_{Con}^{NL+M}$.},
and (5) TOWER-based pre-training of encoder and decoder~\footnote{A 2D U-Net was pre-trained via loss function $\mathcal L_{Con}^{NL+M}+\mathcal L_{Gen}^{NL+M}$.}.
\begin{figure}[htb]
    \centering
    %\vspace{-0.7cm}
    \centerline{\includegraphics[width=.5\linewidth]{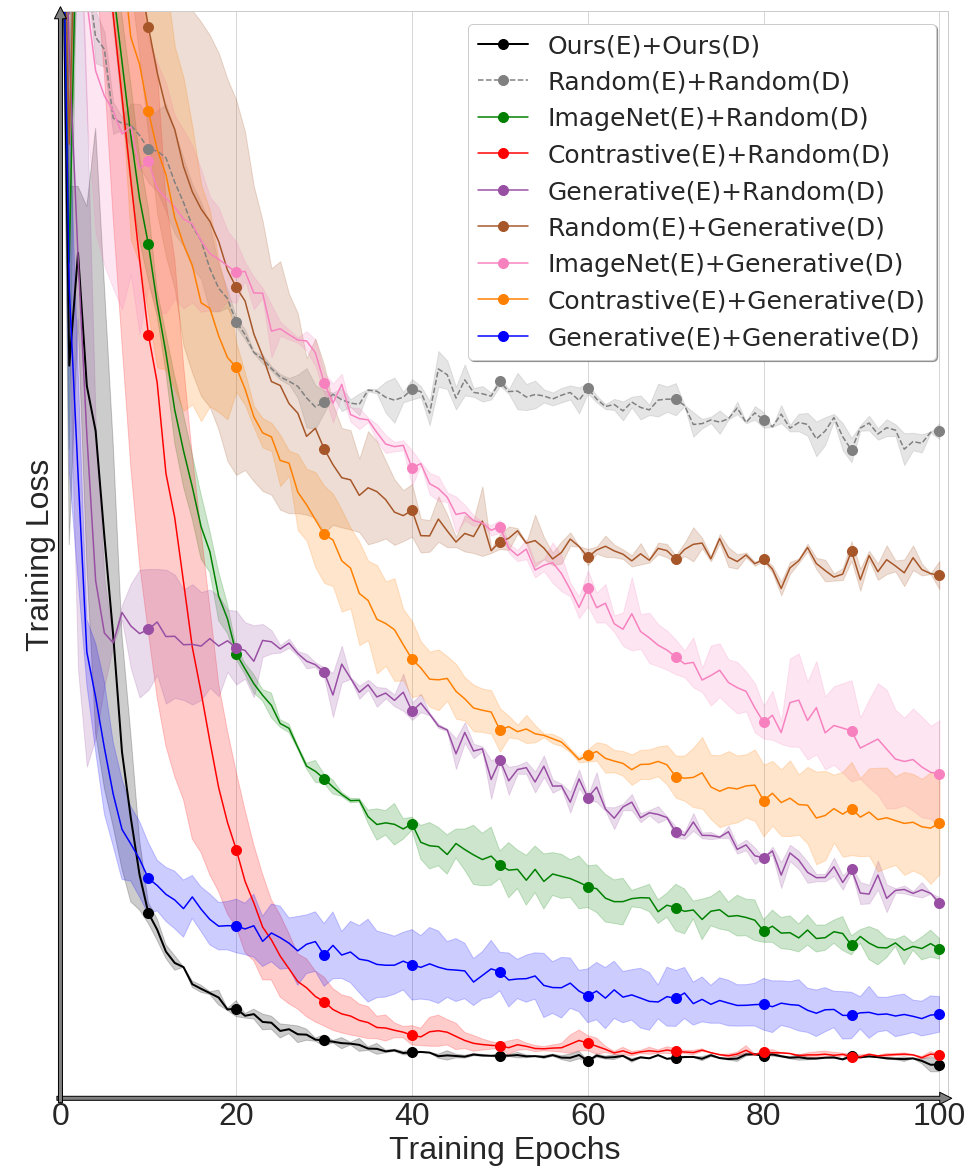}}
    \caption{
    The fine-tuning curves for blood vessel segmentation with different pre-training methods on the fundoscopic images from the DRIVE dataset, where (E) denoted a pre-trained encoder while (D) denoted a pre-trained decoder.}
    %\vspace{-0.5cm}
    \label{fig:finetuning}
\end{figure}

Fig.\ref{fig:finetuning} reported the training loss of the above encoder-decoder-oriented pre-training schemes on the fundoscopic images.
The proposed correlated optimization between the encoder and decoder, i.e., Ours(E)+Ours(D), achieved the best performance and the fastest convergence among all listed methods, demonstrating that TOWER unlocked the power of generative and contrastive learning and provided a significant representation for high-quality segmentation (Issue \#2). 

In contrast, the fine-tuning process of the generative learning-based pre-training method ([Generative(E)+ Generative(D)]) was sub-optimal in segmentation.
Besides, the convergence of the neural network initialized by the classic contrastive learning method was hindered by the random initialization of the decoder, where the convergence speed of [Contrastive(E)+Random(D)] was slower than [TOWER(E)+TOWER(D)], demonstrating the significance of initializing the decoder for contrastive learning in dense pixel prediction tasks. 

Note that when the encoder and decoder were initialized by different pre-training methods, it did not lead to improvements, but rather hindered performance and convergence, e.g., [Contrastive(E)+Generative(D)] required more training epochs to obtain comparable performance to [Contrastive(E)+Random(D)]. The same was true for [ImageNet(E)+Generative(D)]. In contrast, TOWER overcame this pitfall (Issue \#2) and provided semantic-consistent initialization with better performance and faster convergence.

\section{Discussions}\label{ssec:discussion}
This section discussed the potential of applying TOWER to other medical image analysis tasks, including CT images, X-ray images, and ultrasound images. The overview of these datasets was detailed in Table~\ref{tab:datasets}.
\begin{table*}[tb]
% \vspace{-0.7cm}
\caption{An overview of the used datasets in this section to explore the possibility of applying TOWER to other medical image analysis tasks.}
\centering
\resizebox{\linewidth}{!}{
\begin{tabular}{llllll}
\hline
Abbr.$\dagger$    &Dataset    &Task  &Modality&Input size &Scale\\\hline
LXS &Montgomery~\cite{6616679} &Lung segmentation &X-ray &224$\times$224 &138\\\hline
TXC &Shenzhen~\cite{6616679} &Binary-class (2) tuberculosis classification  &X-ray &224$\times$224& 662\\\hline
DXC&CheXpert~\cite{irvin2019chexpert} &Multi-label (5) Binary-class (2) thorax diseases classification&X-ray &224$\times$224 &224,316\\ \hline
BUC &BreastMNIST~\cite{al2020dataset}&Binary-class (2) malignant Classification & Ultrasound &28$\times$28 &780\\\hline
OCC &OrganMNIST~\cite{bilic2019liver} &Multi-label (11) abdominal organ classification&CT &28$\times$28 & 58,851\\\hline
LCS &LiTS~\cite{bilic2019liver} &Liver segmentation &CT &512$\times$512 &130 volumes\\\hline
\multicolumn{6}{p{1.1\linewidth}}{$^\dagger$Abbreviation: the first letter denoted the object of interest, i.e., D denoted thorax diseases, L denoted liver, B denoted Breast, O denoted abdominal organs, and T denoted tuberculosis. The second letter denoted the modality, i.e., X denoted X-ray, U denoted Ultrasound, and C denoted CT. The last letter denoted the task, i.e., C denoted classification, and S denoted segmentation.}\\
% \multicolumn{6}{p{0.96\linewidth}}{$^{**}$ Gaussian units are the same as cg emu for magnetostatics; Mx }
\end{tabular}}
% \vspace{-0.5cm}
\label{tab:datasets}
\end{table*}

Four metrics were highlighted to evaluate TOWER's capability on these tasks, including: (1) the optimal masking ratios in masked reconstruction proxy tasks, (2) the superior performance on various medical image recognition, (3) the distribution of extracted representations on medical image recognition, and (4) the label efficiency via semi-supervised experiments under different label fractions.

\begin{figure}[htb]
    \centering
    %\vspace{-0.7cm}
    \subfigure[Apply the proposed nonlinear translation proxy task.]{% on Shenzhen X-ray dataset
        \begin{minipage}{0.5\textwidth}
        \centerline{\includegraphics[width=\textwidth]{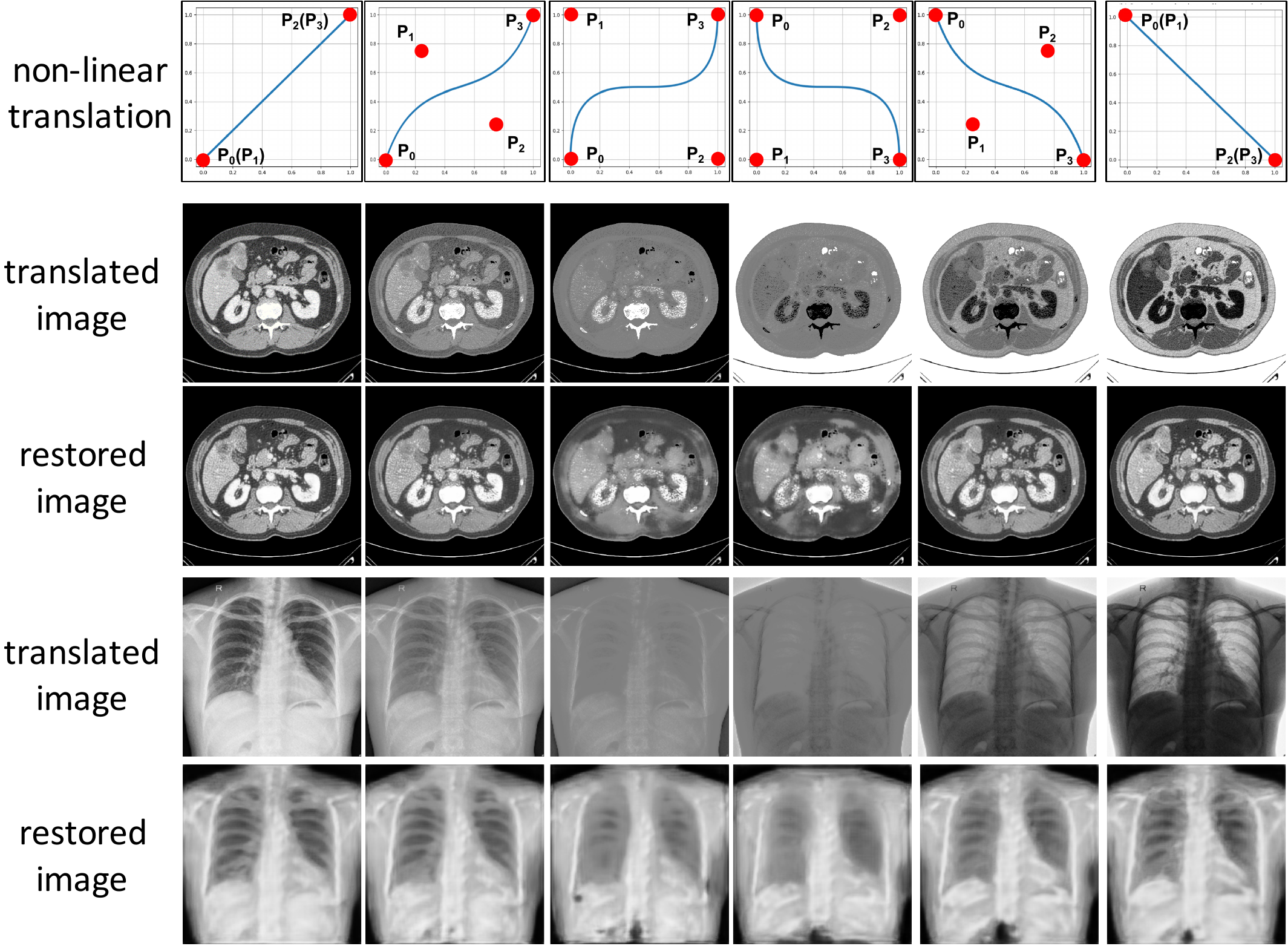}%
        }
    \end{minipage}
    }
    \subfigure[Apply the proposed masked reconstruction proxy task.]{% on Montgomery X-ray dataset
        \begin{minipage}{0.45\textwidth}
        \centerline{\includegraphics[width=\textwidth]{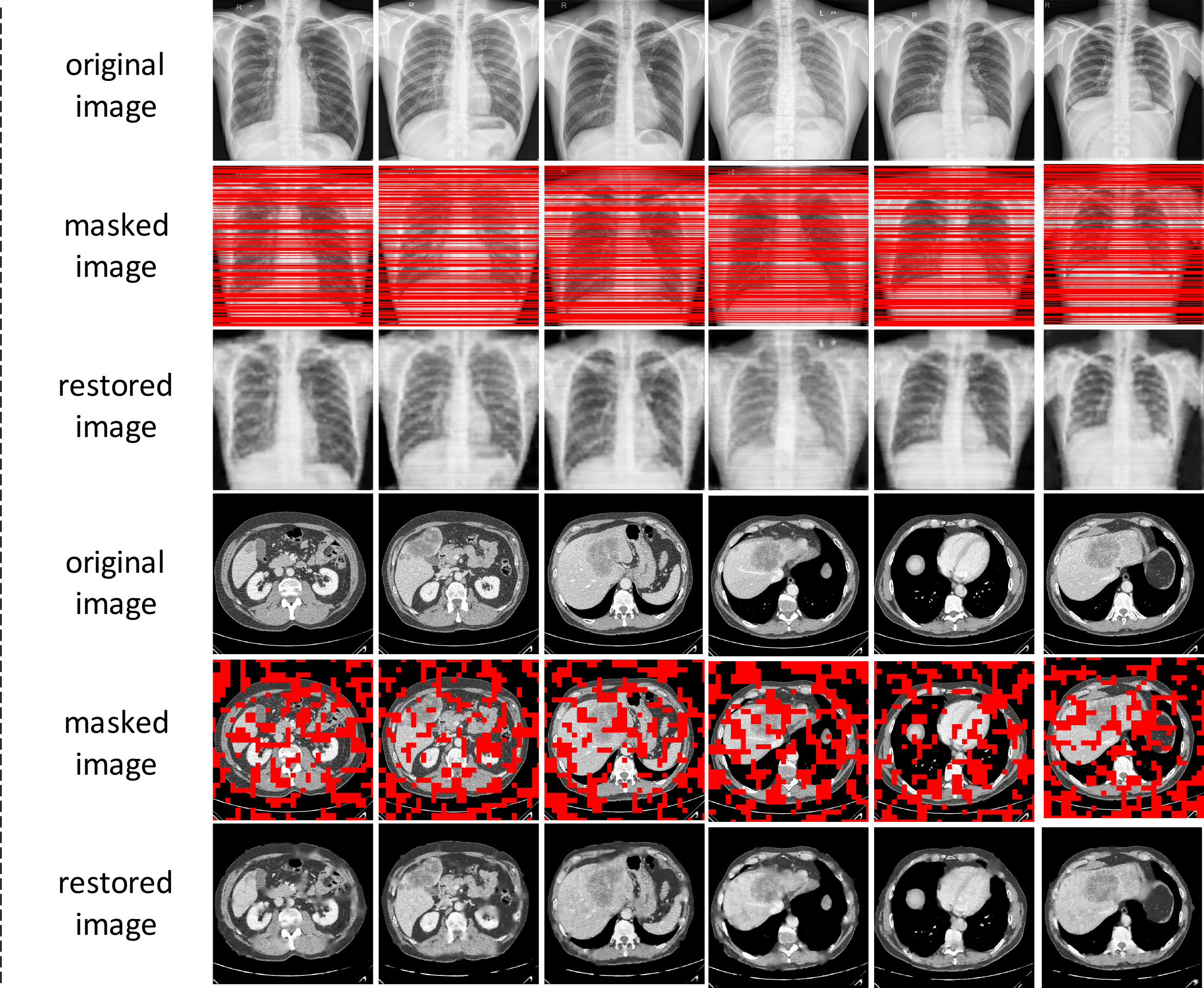}%
        }
        \end{minipage}
    }
    \caption{ 
    Illustrations of the proposed reconstructive proxy tasks on X-ray and CT images.}
    %\vspace{-0.5cm}
    \label{fig:ct-xray}
\end{figure}

The nonlinear translation proxy task was used to enrich the stylistic diversity of all images. Besides, for X-ray images, the stripe-wise masks were designed to reconstruct the texture information of the bones distributed in the stripe region. For CT images, the block-wise masks were designed to reconstruct the distribution of abdominal organs. The above transformations were shown in Fig.\ref{fig:ct-xray}. For ultrasound images, block-wise masks were applied. 

\subsection{The superior performance on various medical image recognition}\label{sssec:comparison}
\begin{table}[tb]
  \centering
  % \vspace{-0.5cm}
   \caption{TOWER outperformed state-of-the-art self-supervised pre-training methods and ImageNet initialization on diverse downstream tasks. Results style: \textbf{best}, \underline{methods that outperformed ImageNet-based initialization}.
%   We also report the p-values between the best and the second best results for each ratio to demonstrate the significance of TOWER.
   }%\vspace{-0.2cm}
% \resizebox{\linewidth}{!}{
    \begin{tabular}{c|c|c|c|c}
    \hline
    Pre-training  & \multicolumn{2}{c|}{Classification tasks (AUC)} & \multicolumn{2}{c}{Segmentation tasks (Dice)} \\ \cline{2-5} 
methods  & TXC & DXC & LCS$\ddagger$ & LXS \\ \hline
Random init.  & 89.03$\pm$1.82 & 86.62$\pm$0.46 & 92.75$\pm$0.57 & 97.55$\pm$0.36 \\
ImageNet init.   & 95.62$\pm$0.63 & 87.10$\pm$0.36 &  94.19$\pm$0.18 &  98.19$\pm$0.13  \\ \hline
InsDis\cite{wu2018unsupervised} &  94.81$\pm$0.73 & \underline{87.21$\pm$0.36} & \underline{94.37$\pm$0.13} & \underline{98.25$\pm$0.03}  \\
Model Genesis\cite{zhou2019models} &  \underline{95.91$\pm$0.63} & \underline{87.79$\pm$0.47}$^{\star\star\star}$ & \underline{94.24$\pm$0.22} &  \underline{98.43$\pm$0.12}  \\
CMC\cite{tian2020contrastive} &  94.93$\pm$1.18 & \underline{87.46$\pm$0.46} & \underline{94.27$\pm$0.26} & \underline{98.49$\pm$0.11}$^{\star\star\star}$  \\
MoCo-v1\cite{he2020momentum} &   94.54$\pm$0.42 & 86.98$\pm$0.11 & 93.87$\pm$0.72 & 98.08$\pm$0.14  \\
PIRL\cite{misra2020self} &  93.34$\pm$2.72 & 86.79$\pm$0.35 & 94.05$\pm$0.21 & 98.02$\pm$0.11 \\
SimCLR-v1\cite{chen2020simple} &   94.45$\pm$0.76 & \underline{87.66$\pm$0.14} & 93.54$\pm$0.18 & 98.19$\pm$0.10  \\
MoCo-v2\cite{chen2020improved} &  95.57$\pm$0.90 & 86.94$\pm$0.20 & 93.88$\pm$0.21 & 97.79$\pm$0.50  \\
SimCLR-v2\cite{NEURIPS2020_fcbc95cc} &   95.29$\pm$0.93 & 86.86$\pm$0.37 & 93.85$\pm$0.17 & 98.16$\pm$0.20  \\
SeLa-v2\cite{YM.2020Self-labelling,NEURIPS2020_70feb62b} &   \underline{96.23$\pm$0.81} & \underline{87.24$\pm$0.29} & 93.98$\pm$0.21 & \underline{98.28$\pm$0.04}  \\
InfoMin\cite{NEURIPS2020_4c2e5eaa} &  95.02$\pm$1.40 & 86.67$\pm$0.10 & 94.26$\pm$0.17 & 97.94$\pm$0.16  \\
BYOL\cite{NEURIPS2020_f3ada80d} &  94.69$\pm$0.78 & 87.09$\pm$0.40 & 93.76$\pm$0.25 & \underline{98.20$\pm$0.08}  \\
DeepCluster-v2\cite{caron2018deep}&   \underline{96.09$\pm$0.68} & 87.01$\pm$0.19 & 93.84$\pm$0.25 & \underline{98.24$\pm$0.05}\\
SwAV\cite{NEURIPS2020_70feb62b} &   \underline{95.72$\pm$0.50} & 87.06$\pm$0.50 & 93.63$\pm$0.31 & \underline{98.28$\pm$0.05} \\
PCL-v1\cite{li2021prototypical} &  95.15$\pm$0.53 & 86.90$\pm$0.25 & \underline{94.55$\pm$0.09}$^{\star}$ & \underline{98.25$\pm$0.05}  \\
PCL-v2\cite{li2021prototypical} &  95.45$\pm$0.62 & \underline{87.27$\pm$0.19} & 93.95$\pm$0.20 & \underline{98.26$\pm$0.06} \\
Barlow Twins\cite{pmlr-v139-zbontar21a} &  94.50$\pm$0.88 & \underline{87.25$\pm$0.27} & 93.86$\pm$0.07 & \underline{98.23$\pm$0.05} \\ 
CAiD\cite{taher2022caid} & / & \underline{87.44$\pm$0.33} & / & 98.19$\pm$0.08  \\
DiRA\cite{haghighi2022dira} & / & \underline{87.59$\pm$0.28} & / & \underline{98.24$\pm$0.09}  \\\hline
TOWER (Ours) &  \textbf{96.53$\pm$0.50} & \textbf{88.89$\pm$0.33} & \textbf{94.66$\pm$0.05} & \textbf{98.95$\pm$0.08} \\ \hline
    \end{tabular}%
    
    \resizebox{\linewidth}{!}{
    \begin{tabular}{ccccc}
\multicolumn{5}{l}{$\ddagger$ Reproduced by ourselves under the same protocol.} \\
\multicolumn{5}{l}{/ denoted not reported in the original paper.} \\
\multicolumn{5}{p{\linewidth}}{$\star$ Statistical analysis between the \textbf{SECOND-BEST} method and our proposed \textbf{TOWER} was also conducted in each target task, where $^{\star}$ denoted TOWER significantly outperformed the second-best method with p-value\textless0.005. $^{\star\star}$ denoted p-value\textless0.001. $^{\star\star\star}$ denoted p-value\textless0.0001.}
    \end{tabular}}
  \label{tab:sotaresults}%
    %\vspace{-0.5cm}
\end{table}%

TOWER achieved the best performance among the state-of-the-art methods in recognizing fundoscopic images (see Table~\ref{tab:sotaresults-vfs}). This section compared TOWER with the these methods on various medical image recognition tasks, including lung segmentation from X-ray images \footnote{\url{www.kaggle.com/datasets/kmader/pulmonary-chest-xray-abnormalities}} (Montgomery~\cite{6616679}), liver segmentation from CT images \footnote{\url{https://competitions.codalab.org/competitions/17094}} (LiTS~\cite{bilic2019liver}), tuberculosis classification from X-ray images \footnote{\url{http://openi.nlm.nih.gov/imgs/collections/ChinaSet\_AllFiles.zip}} (Shenzhen~\cite{6616679}), and thorax diseases classification from X-ray images \footnote{\url{https://stanfordmlgroup.github.io/competitions/chexpert/}}) (CheXpert~\cite{irvin2019chexpert}). 

Table~\ref{tab:sotaresults} reported that TOWER successfully enhanced contrastive learning by enriching the insufficient diversity of other medical images.
Table~\ref{tab:sotaresults} also demonstrated that TOWER unleashed the power of 2D pre-training for medical image recognition, excelled in obtaining the significant representations demanded by a high-quality segmentation, and could be a better alternative to ImageNet-based pre-training in a wide range of medical image recognition tasks.
Table~\ref{tab:ablation} conducted the ablation studies on the BUC and OCC classification tasks. The same conclusions as in Section ~\ref{ssec:ablation} can also be obtained, demonstrating the effectiveness of TOWER on other medical images.
% This study attributed this performance superiority to the synergistic effect of contrastive and generative learning with a priori knowledge.
\begin{table}[tbp]
  \centering
  %\vspace{-0.5cm}
  \caption{Performance of the proposed self-supervised pre-training components on various downstream tasks. Results style: \textbf{best}.
%   The significance test is reported in Table \ref{tab:p-value} and \ref{tab:p-value-2}. 
%   Symbol * indicates statistically significant improvement given by a Wilcoxon signed-rank test with $p\leq0.05$.
  }%\vspace{-0.4cm}
	\begin{center}
	% \resizebox{\linewidth}{!}{
    \begin{tabular}{c|cccc|cc}
    \hline
%     \multirow{1}[3]{*}{Methods} & \multirow{1}[2]{*}{$\mathcal L_{Con}^{moco}$} & \multirow{1}[3]{*}{$\mathcal L_{Gen}^{NL}$} & \multirow{1}[2]{*}{$\mathcal L_{Gen}^{M}$} & \multirow{1}[3]{*}{$\mathcal L_{Con}^{NL+M}$} & PathMNIST & BreastMNIST & DermaMNIST & RetinaMNIST & OrganMNIST & DRIVE \\
% \cline{6-11}          &  &  &  &  & \multicolumn{6}{c}{AUC or Dice (\%) $\uparrow$} \\
%     \hline
% \cline{6-11}          &  &  &  &  & \multicolumn{6}{c}{AUC or Dice (\%) $\uparrow$} \\
%     \hline
    \multirow{2}[2]{*}{Methods} & \multirow{2}[2]{*}{$\mathcal L_{Con}^{moco}$} & \multirow{2}[2]{*}{$\mathcal L_{Gen}^{NL}$} & \multirow{2}[2]{*}{$\mathcal L_{Gen}^{M}$} & \multirow{2}[2]{*}{$\mathcal L_{Con}^{NL+M}$} & \multicolumn{2}{c}{Classification AUC ($\%$)}   \\
    &  &  &  &  & BUC   & OCC   \\
\hline
    Random &  &  &  &  & 83.42$\pm$2.77 & 98.92$\pm$0.17  \\
    ImageNet &  &  &  &  & 86.48$\pm$2.73 & 99.52$\pm$0.11 \\
    MoCo  & $\checkmark$ &  &  &   & 87.80$\pm$2.49  &  99.51$\pm$0.06  \\
    \hline
    \multicolumn{1}{c|}{\multirow{3}[5]{*}{TOWER}} &  & $\checkmark$ &  &  & 86.23$\pm$3.65&  99.49$\pm$0.06  \\
          &  &  & $\checkmark$ &   & 85.58$\pm$2.99  & 99.49$\pm$0.07  \\
          &  & $\checkmark$ & $\checkmark$ &   & 87.25$\pm$1.77  & 99.51$\pm$0.06  \\
          & &  &  & $\checkmark$   & 88.21$\pm$2.90   & 99.57$\pm$0.06 \\
          &  &$\checkmark$ & $\checkmark$ & $\checkmark$  & \textbf{88.40$\pm$1.04}  & \textbf{99.60$\pm$0.04}  \\
    \hline
    \end{tabular}%
    \end{center}
    %\vspace{-0.8cm}
  \label{tab:ablation}%
\end{table}%
\begin{figure*}[htb]
    \centering
    \subfigure[Performance of different ratios of strip-wise masks on X-ray images.]{% on Shenzhen X-ray dataset
    \vspace{-0.3cm}
        \begin{minipage}{0.32\textwidth}
        \centerline{\includegraphics[width=\textwidth]{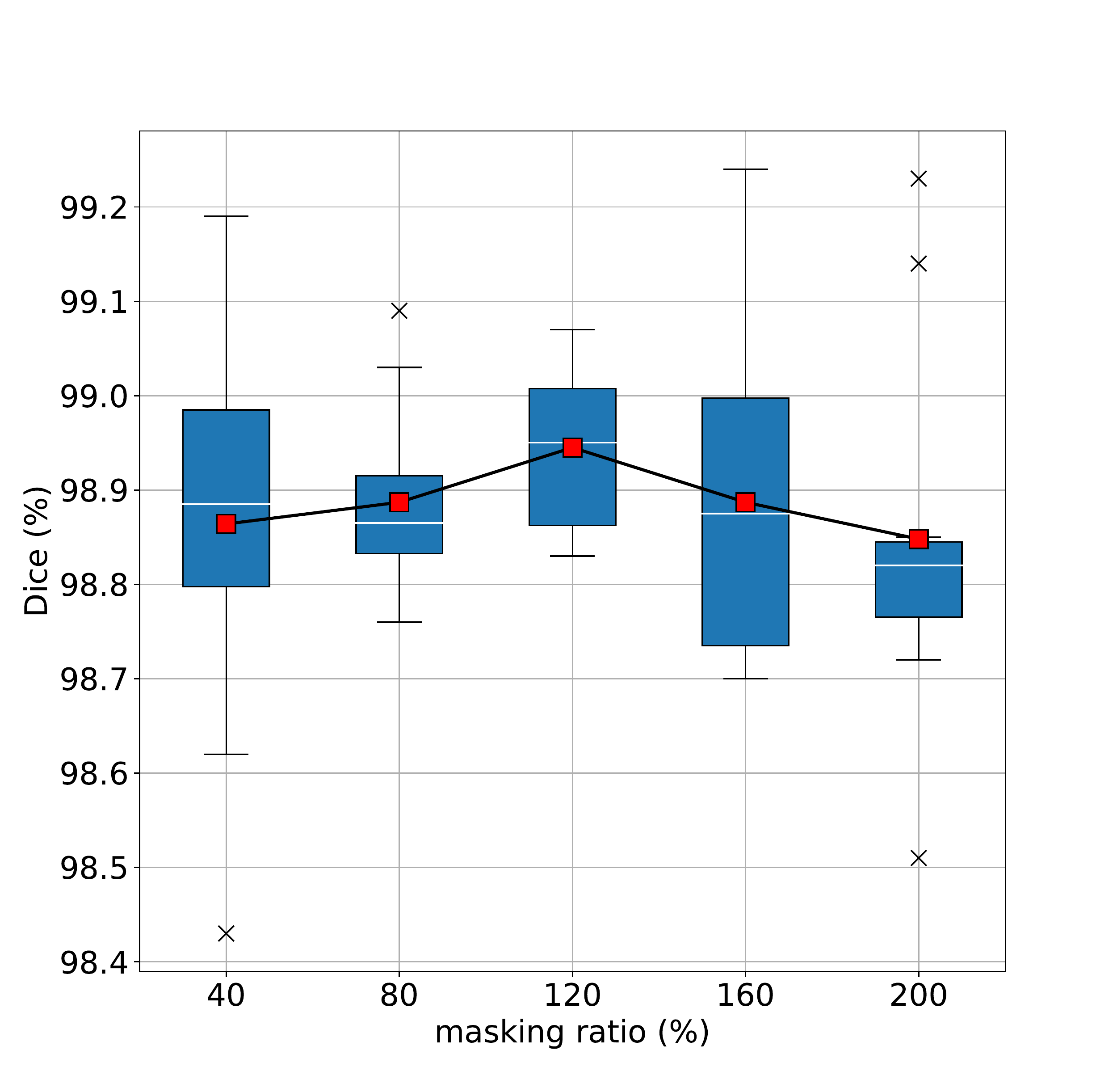}%
        }
    \end{minipage}
    }
    \subfigure[Performance of different ratios of block-wise masks on CT images.]{% on Shenzhen X-ray dataset
        \begin{minipage}{0.32\textwidth}
    % %\vspace{-0.6cm}
        \centerline{\includegraphics[width=\textwidth]{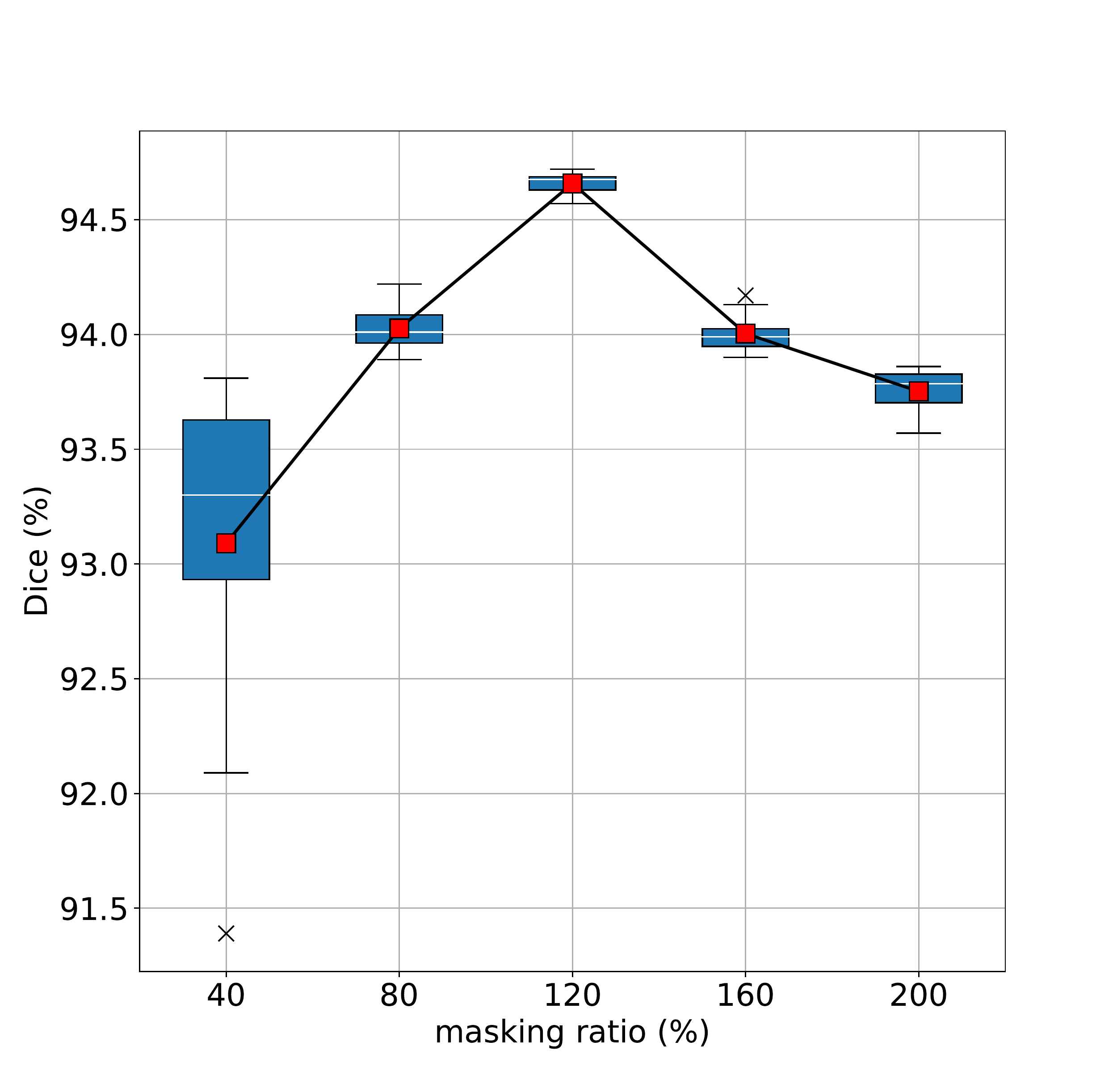}%
        }
    \end{minipage}
    }
    \subfigure[Performance of different number of rays in the rays-wise masks on fundoscopic images.]{% on Montgomery X-ray dataset
        \begin{minipage}{0.32\textwidth}
    % %\vspace{-0.6cm}
        \centerline{\includegraphics[width=\textwidth]{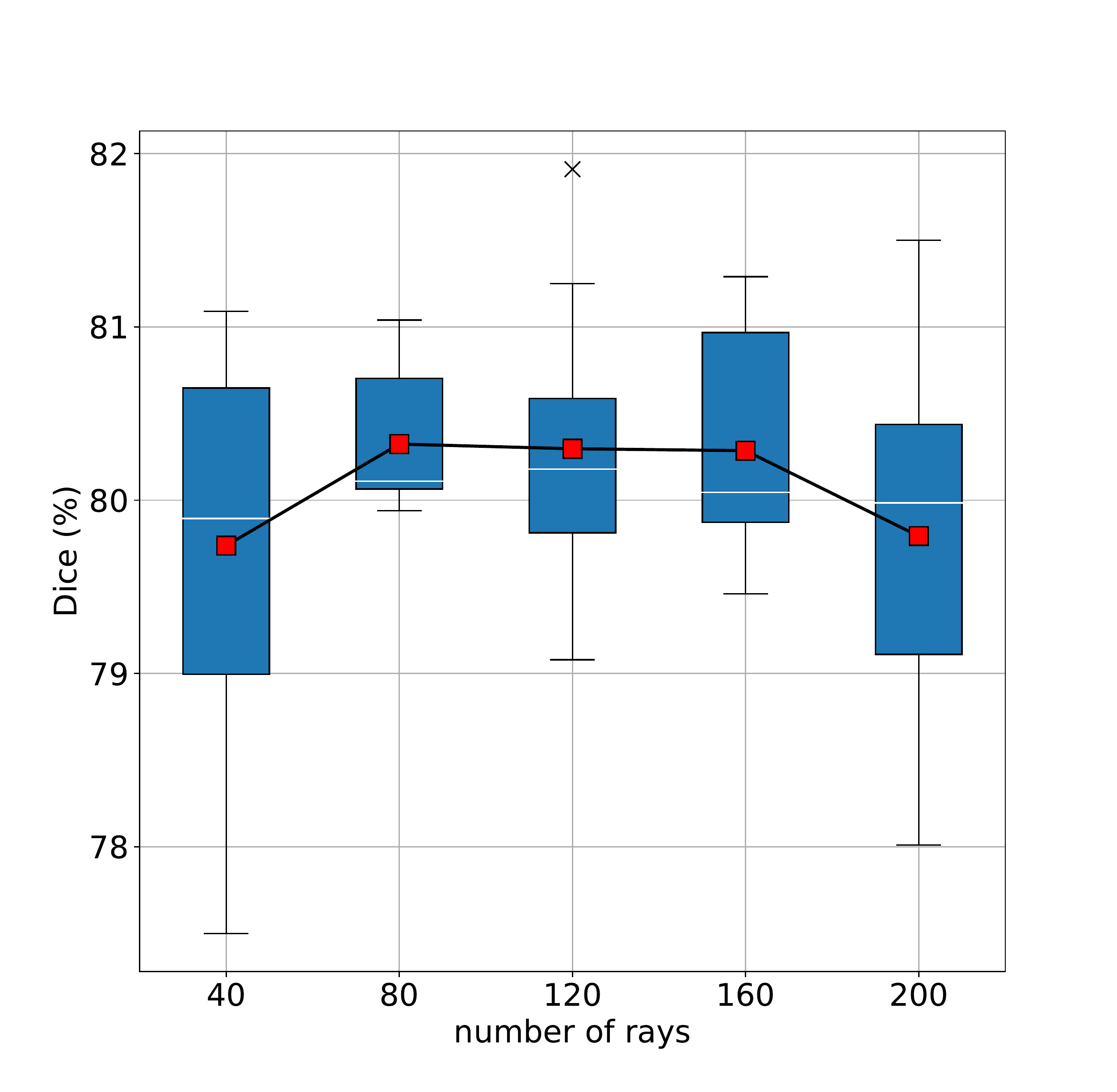}%
        }
        \end{minipage}
    }
    % %\vspace{-0.6cm}
    \caption{An experiment of the impact of different masking ratios on downstream tasks. The boxplots showed fine-tuning results for recognizing medical images with different modalities and at different masking ratios. The red dot indicated the mean value across ten independent trials.}
    %\vspace{-0.5cm}
    \label{fig:ratio}
\end{figure*}
\begin{figure*}[tb]
    \centering
    %\vspace{-0.7cm}
    \subfigure[Apply MoCo to the MNIST multi-labels classification task of handwritten digits.]{% on Shenzhen X-ray dataset
        \begin{minipage}{0.32\textwidth}
        \centerline{\includegraphics[width=.8\textwidth]{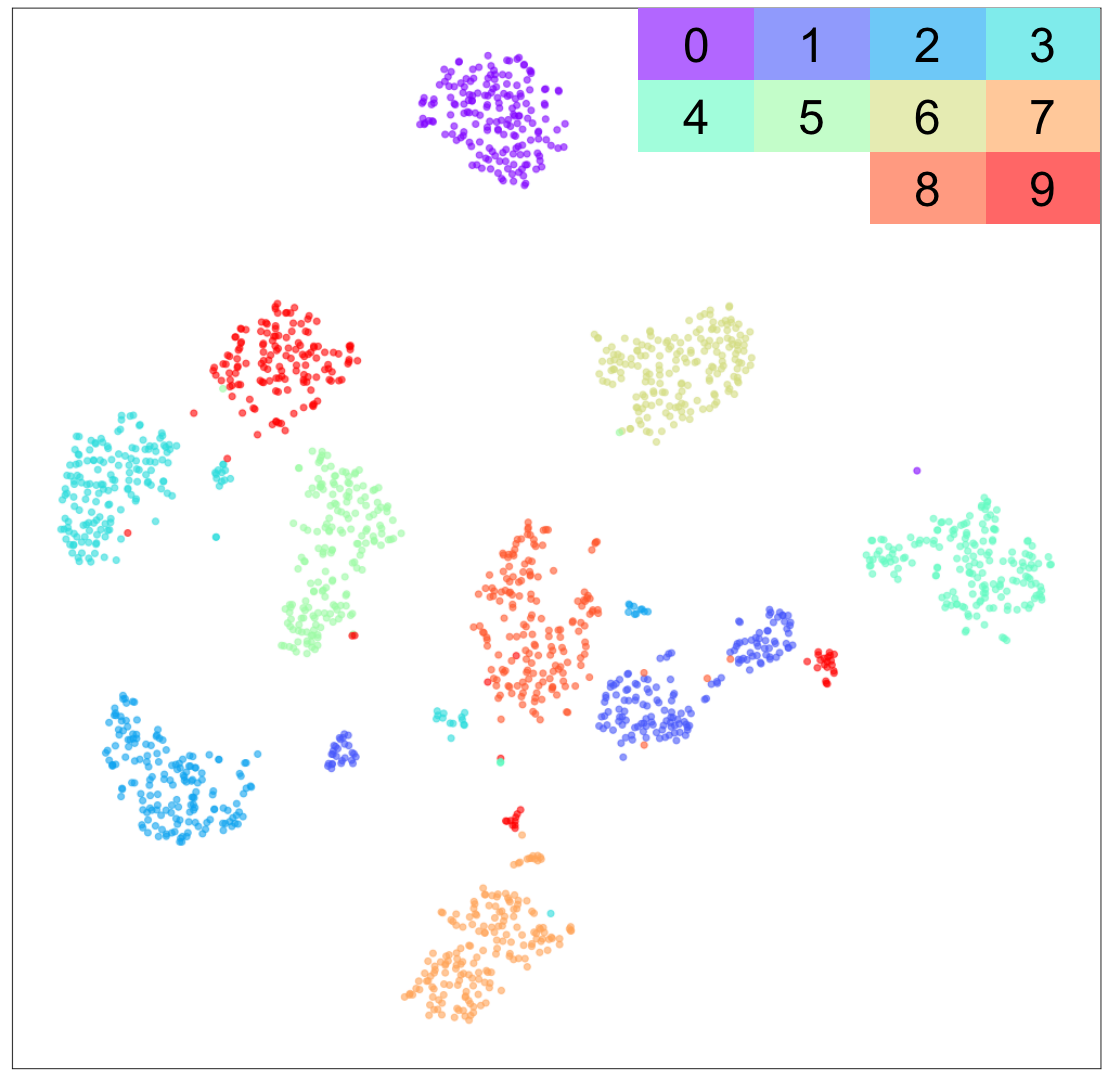}%
    }\end{minipage}
    }
    \subfigure[Apply MoCo to classify multi-labels abdominal organs on OrganMNIST dataset.]{% on Montgomery X-ray dataset
        \begin{minipage}{0.32\textwidth}
        \centerline{\includegraphics[width=.8\textwidth]{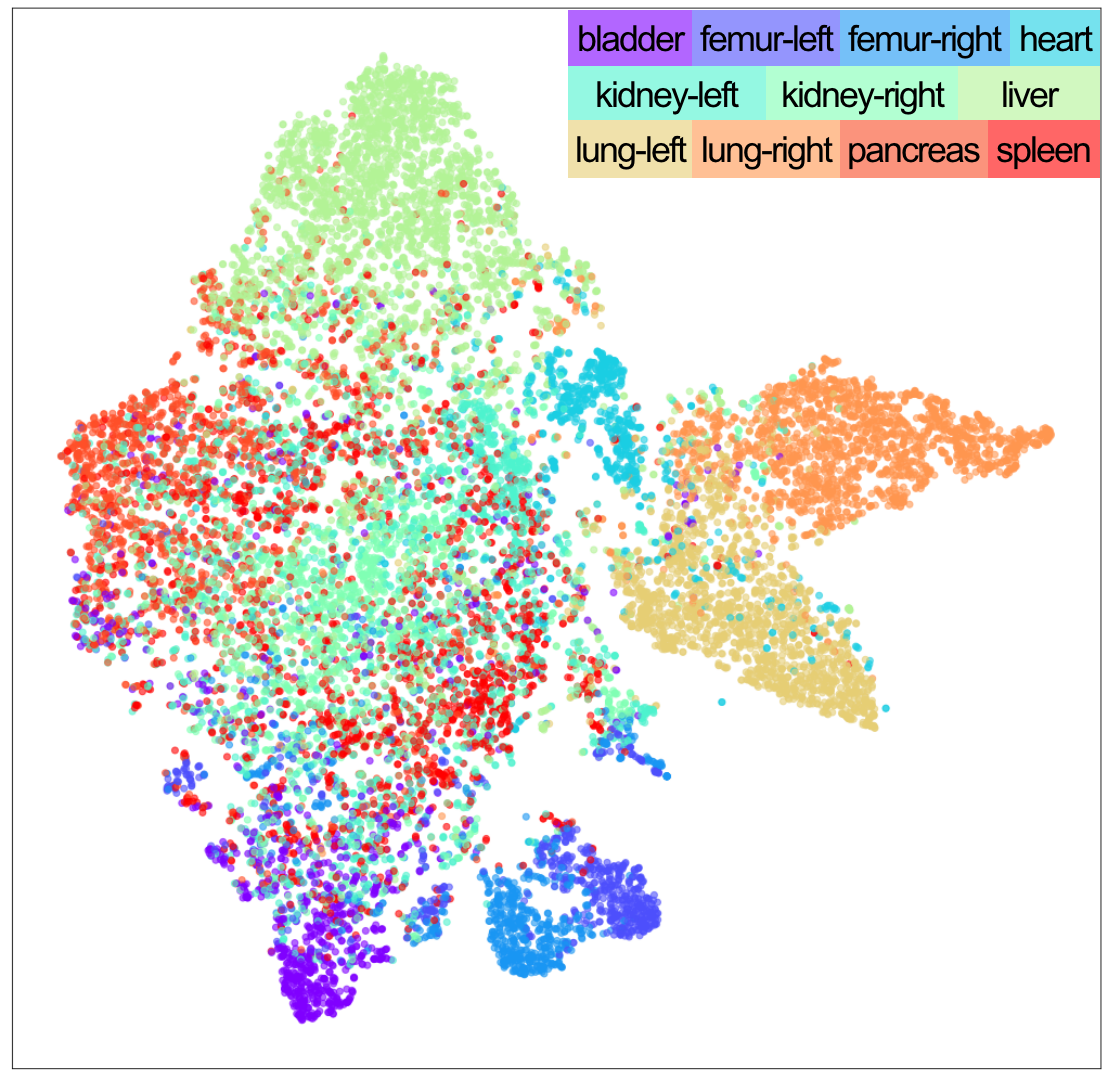}%
        }\end{minipage}
    }
    %\vspace{-0.2cm}
    \subfigure[Apply TOWER to classify multi-labels abdominal organs on OrganMNIST dataset.]{% on Shenzhen X-ray dataset
        \begin{minipage}{0.32\textwidth}
        \centerline{\includegraphics[width=.8\textwidth]{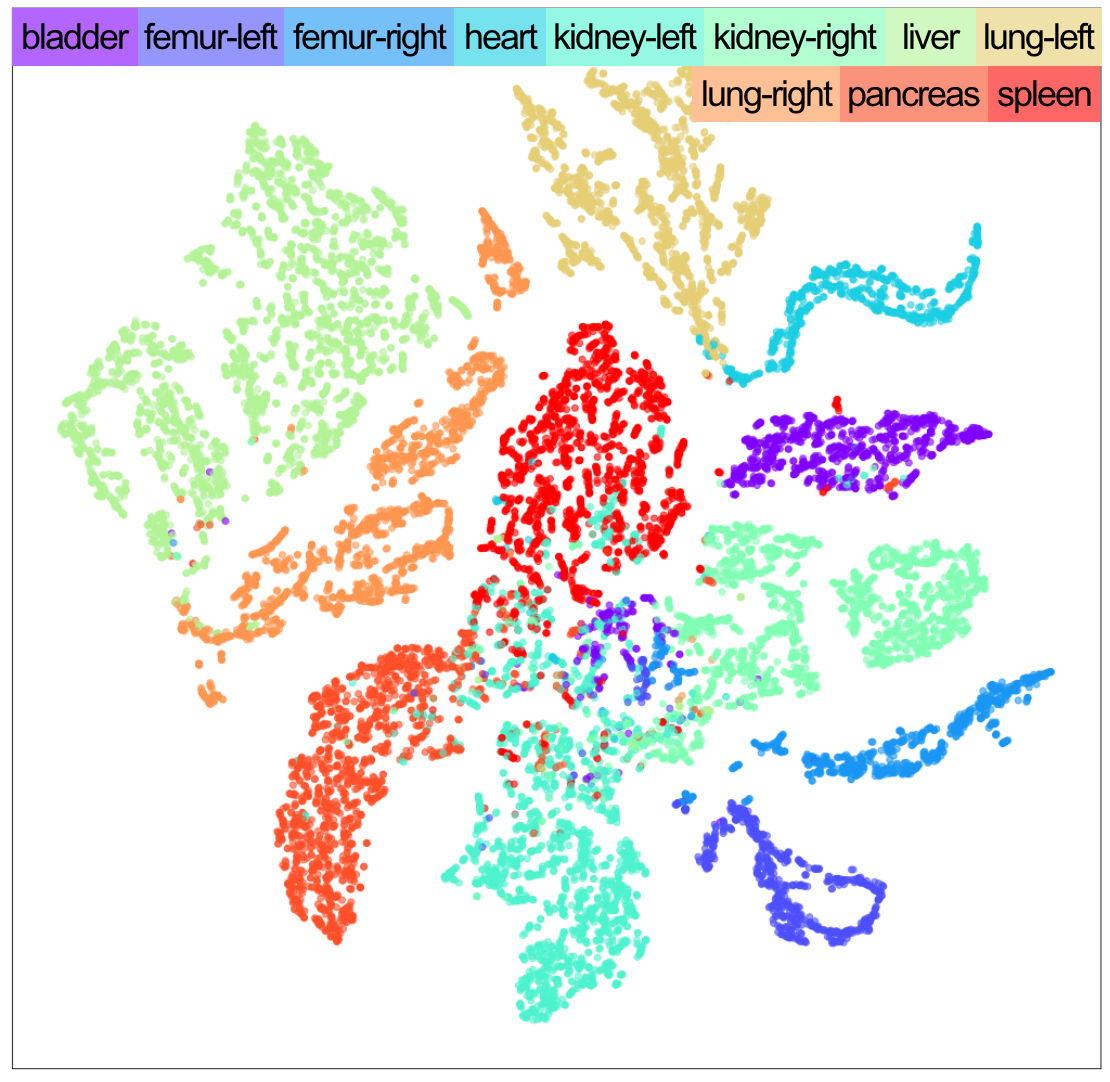}%
    }\end{minipage}
    }
    \caption{ 
    Illustration of the extracted representations on different tasks (zoom in for more details). Fig.(a) was trained on MNIST. Fig.(b) and Fig.(c) were trained on OCC.}
    %\vspace{-0.5cm}
    \label{fig:visualization}
\end{figure*}
\subsection{The optimal masking ratios for reconstruction}\label{sssec:ratio}
This section explored the best masking ratios on the three most common medical image modalities. Fig.\ref{fig:ratio} showed the performance of different masking ratios. The optimal ratio varied for different kinds of masks, where the best number of rays in the rays-wise mask was 80 for retinal images. In the block-wise mask for CT images and stripe-wise mask for X-ray images, the best masking ratios were 50\%.

This finding was consistent with the understanding of the mask reconstruction proxy task. A low mask ratio led to a simple proxy task, and a high mask ratio led to insufficient image information. These improper mask ratios resulted in difficulties for the neural network to learn significant representations.

\subsection{Visualizing the extracted robust representations}\label{sssec:visualize}
In order to explore the impact of the proposed restorative proxy tasks on contrastive learning, the representations extracted by the encoder were visualized via t-SNE~\cite{van2008visualizing}.
The classic contrastive learning method (MoCo) was selected as the baseline. This study conducted handwritten digits classification on MNIST~\cite{lecun1998gradient} and multi-organs classification on OCC to evaluate the MoCo and TOWER.

Fig.\ref{fig:visualization}(a)(b) showed that MoCo could cluster representations of natural images correctly but failed in the medical image classification task. In contrast, Fig.\ref{fig:visualization}(c) showed that the proposed knowledge-based restorative proxy tasks enriched the sample diversity of medical images, thus successfully constructing robust positive/negative pairs and correctly clustering the unlabelled representations.

\subsection{Label-efficient learning with few labelled samples}\label{sssec:semi}
\begin{figure*}[tb]
    %\vspace{-0.5cm}
    \centerline{\includegraphics[width=\textwidth]{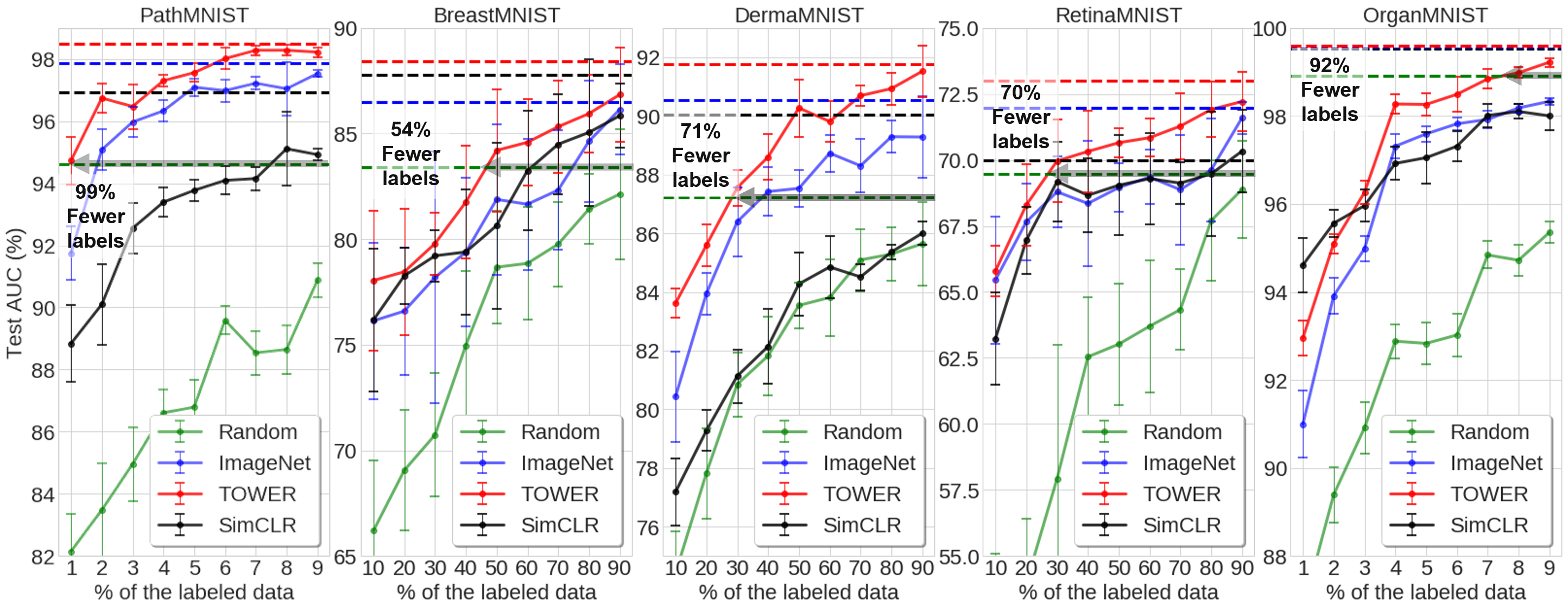}%
    }
    \caption{Results of the semi-supervised experiments under different label fractions. The horizontal red, blue, black, and green dash lines indicated the performance (AUC) by TOWER, ImageNet, SimCLR v1, and random initialization on the 100\% labelled test dataset, respectively. The gray arrows represented the proportion of labelled data that the TOWER-based method can reduce compared to the random initialization method.}
    %\vspace{-0.5cm}
    \label{fig:semi}
\end{figure*}
In order to explore the label-efficiency of TOWER on different percentages of labelled data, this section conducted semi-supervised experiments on PathMNIST, BreastMNIST, DermaMNIST, RetinaMNIST, and OrganMNIST.

Fig.\ref{fig:semi} displayed the test AUC (\%) of the neural network initialized from random, ImageNet, and TOWER under different label fractions. 
Results demonstrated that TOWER mitigated the lack of annotations and achieved label-efficient representation learning for medical image recognition.
With decreasing amounts of labelled data, TOWER retained a much higher performance on all downstream tasks, especially when few labelled samples were available. Besides, TOWER could be fine-tuned on a few labelled datasets to achieve comparable performance to the full-labelled (100\%) dataset.
Specifically, compared to training from scratch, initialization with TOWER could reduce the annotation cost by 99\%, 54\%, 71\%, 70\%, and 92\% for PathMNIST, BreastMNIST, DermaMNIST, RetinaMNIST, and OrganMNIST, respectively.
Compared to ImageNet-based supervised pre-training, initializing with TOWER could reduce the annotation cost by 94\%, 15\%, 31\%, and 20\% for PathMNIST, BreastMNIST, DermaMNIST, RetinaMNIST, respectively.
Compared to SimCLR v1-based self-supervised pre-training, initializing with TOWER could reduce the annotation cost by 96\%, 52\%, and 70\% for PathMNIST, DermaMNIST, RetinaMNIST, respectively.

%\subsection{Summary}\label{ssec:summary}

Overall, TOWER\footnote{The source code and the pre-trained weights are available at \url{https://github.com/lichen14/TOWER}.} had made significant progresses towards solving the open issues (Section~\ref{sec:introduction}):
(1) Robust representations could be learned in an unsupervised manner from unlabelled biomedical microscopy images by enriching low diversity with a priori knowledge-based proxy tasks;
(2) High-quality segmentation had been enabled by bridging the generative and contrastive learning to co-optimize the encoder and the decoder with semantic-consistency. 

\section{Conclusions}
\label{sec:conclusion}

Aiming at the grand challenges for automated recognition of biomedical microscopy images towards clinical practices, this study developed a knowledge-based learning framework (TOWER). The framework synergizes generative learning and contrastive learning to enhance self-supervised learning towards high-quality initialization of AE neural networks for reliable classification and segmentation tasks.  

TOWER enabled sample space diversification via reconstructive proxy tasks to perform the nonlinear translation and random masked reconstruction based on a priori knowledge from clinic practices. The design supported enhanced representation learning of the representation of annotation-free images. Correlated optimization of the encoder and the decoder had been achieved by bridging contrastive and generative learning to serve the need for semantic segmentation.

Experimental results indicated that: (1) the proposed restorative proxy tasks could enrich the diversity of biomedical microscopy images and enhance contrastive learning with extended sample space; (2) TOWER could bridge generative and contrastive learning as a whole and provide better performance and faster convergence for segmentation by correlated-optimization between the encoder and decoder; (3) TOWER could transfer to other medical image recognition tasks and mitigate the lack of annotations, resulting in label-efficient representation learning for medical image recognition (reduce up to 99\% annotations in pathological classification).

Overall, TOWER significantly outperformed even the latest high-performance counterparts in terms of recognizing unlabelled biomedical microscopy images. It suggested that the key to sustaining reliable recognition of unlabelled medical images lay with appropriate portraying a priori knowledge in optimization.

% Numbered list
% Use the style of numbering in square brackets.
% If nothing is used, default style will be taken.
%\begin{enumerate}[a)]
%\item 
%\item 
%\item 
%\end{enumerate}  

% Unnumbered list
%\begin{itemize}
%\item 
%\item 
%\item 
%\end{itemize}  

% Description list
%\begin{description}
%\item[]
%\item[] 
%\item[] 
%\end{description}  

%% Figure
%\begin{figure}[<options>]
%	\centering
%		\includegraphics[<options>]{}
%	  \caption{}\label{fig1}
%\end{figure}
%
%
%\begin{table}[<options>]
%\caption{}\label{tbl1}
%\begin{tabular*}{\tblwidth}{@{}LL@{}}
%\toprule
%  &  \\ % Table header row
%\midrule
% & \\
% & \\
% & \\
% & \\
%\bottomrule
%\end{tabular*}
%\end{table}
%
%% Uncomment and use as the case may be
%%\begin{theorem} 
%%\end{theorem}
%
%% Uncomment and use as the case may be
%%\begin{lemma} 
%%\end{lemma}
%
%%% The Appendices part is started with the command \appendix;
%%% appendix sections are then done as normal sections
%%% \appendix
%
%\section{}\label{}
%
%% To print the credit authorship contribution details
%\printcredits
%
%%% Loading bibliography style file
%%\bibliographystyle{model1-num-names}
%\bibliographystyle{cas-model2-names}
%
%% Loading bibliography database
% \bibliographystyle{elsarticle-num}
\bibliographystyle{cas-model2-names}
\bibliography{cas-refs.bib}
\end{document}